\documentclass{article}

\usepackage{arxiv}

\usepackage[utf8]{inputenc} % allow utf-8 input
\usepackage[T1]{fontenc}    % use 8-bit T1 fonts
\usepackage{hyperref}       % hyperlinks
\usepackage{url}            % simple URL typesetting
\usepackage{booktabs}       % professional-quality tables
\usepackage{amsfonts}       % blackboard math symbols
\usepackage{nicefrac}       % compact symbols for 1/2, etc.
\usepackage{microtype}      % microtypography
\usepackage{lipsum}		% Can be removed after putting your text content
\usepackage{graphicx}
\usepackage[numbers]{natbib}
\usepackage{doi}

\usepackage{subfig}
\usepackage[section]{placeins}

\usepackage{ntheorem,hyperref}
\theoremseparator{:}
\newtheorem{hypx}{Hypothesis}
\newenvironment{hyp}{\hypx[H\textsubscript{\thehypx}]}{\endhypx}
\newcommand{\hypref}[1]{\textup{H\textsubscript{\ref{#1}}}}

\usepackage{booktabs}
\usepackage{times}
\usepackage{cite}
\usepackage{xcolor}
\usepackage{url}
\usepackage{multirow}
\usepackage{adjustbox}
\newcommand{\wrt}{{\it w.r.t. }}    % with respect to
\newcommand{\eg}{\emph{e.g.}, }     % for example
\newcommand{\ie}{\emph{i.e.}, }     % that is
\newcommand\etc{\emph{etc.}}

\title{The MAMe Dataset: On the relevance of High Resolution and Variable Shape image properties}

\author{Ferran Parés \\
	Barcelona Supercomputing Center (BSC) \\
	\texttt{ferran.pares@bsc.es} \\
	\And
	Anna Arias-Duart \\
	Barcelona Supercomputing Center (BSC) \\
	\texttt{anna.ariasduart@bsc.es} \\
	\And
	Dario Garcia-Gasulla \\
	Barcelona Supercomputing Center (BSC) \\
	\texttt{dario.garcia@bsc.es} \\
	\And
	Gema Campo-Francés \\
	Conservació-Restauració Universitat de Barcelona (UB) \\
	\And
	Nina Viladrich \\
	Conservació-Restauració Universitat de Barcelona (UB) \\
	\And
	Eduard Ayguadé \\
	Barcelona Supercomputing Center (BSC) \\
	Universitat Politècnica de Catalunya (UPC) \\
	\And
	Jessús Labarta \\
	Barcelona Supercomputing Center (BSC) \\
	Universitat Politècnica de Catalunya (UPC) \\
}
\date{}

\begin{document}

\maketitle

\begin{abstract}
In the image classification task, the most common approach is to resize all images in a dataset to a unique shape, while reducing their precision to a size which facilitates experimentation at scale. This practice has benefits from a computational perspective, but it entails negative side-effects on performance due to loss of information and image deformation. In this work we introduce the MAMe dataset, an image classification dataset with remarkable high resolution and variable shape properties. The goal of MAMe is to provide a tool for studying the impact of such properties in image classification, while motivating research in the field. The MAMe dataset contains thousands of artworks from three different museums, and proposes a classification task consisting on differentiating between 29 mediums (\ie materials and techniques) supervised by art experts. After reviewing the singularity of MAMe in the context of current image classification tasks, a thorough description of the task is provided, together with dataset statistics. Experiments are conducted to evaluate the impact of using high resolution images, variable shape inputs and both properties at the same time. Results illustrate the positive impact in performance when using high resolution images, while highlighting the lack of solutions to exploit variable shapes. An additional experiment exposes the distinctiveness between the MAMe dataset and the prototypical ImageNet dataset. Finally, the baselines are inspected using explainability methods and expert knowledge, to gain insights on the challenges that remain ahead.
\end{abstract}


\section{Introduction}\label{sec:i}

%% datasets drive research
Challenging problems is what drives AI research. What pushes the field and its applications forward. A prime example of that is the ImageNet dataset, together with the corresponding ILSVRC challenge \citep{russakovsky2015imagenet}. The popularization of this competition revitalized the Neural Networks field, particularly in the context of image processing. The outstanding performance of deep neural networks models in the demanding ILSVRC challenge caught the attention of AI researchers and practitioners around the world, who quickly acknowledged the potential behind the combination of deep nets and large sets of data. As a result, the popularity of the field exploded.

The ImageNet dataset provided an appealing challenge to lure AI researchers, who in turn were able to develop and test new ideas on it. Some of these ideas became powerful principles for the current deep learning (DL) field, such as Inception blocks \citep{szegedy2015going}, residual connections \citep{he2016deep}, dropout regularization \citep{srivastava2014dropout}, ReLU activations \citep{nair2010rectified} and weight initializations \citep{glorot2010understanding, he2015delving}, among others. This amounts for a remarkable set of achievements in a very short time span, and speaks of the contribution of ImageNet to the AI field. That being said, the relevance of the ImageNet image classification challenge today has mostly vanished. The last edition of ILSVRC took place in 2017 \citep{Ilsvrc17last}, and the AI community considers it a solved problem with little margin for improvement (by 2019, 98.2\% top-5 accuracy \citep{xie2019selftraining} was achieved, while human top-5 classification accuracy is thought to be between 88\% and 95\%~\citep{russakovsky2015imagenet}).

The ImageNet challenge is defined around two main types of instances: Man-made objects, and living things. These classes are characterized by large distinctive features which require little attention to detail for their recognition. State-of-the-art performance can be achieved on this kind of tasks after applying heavy deformation on the image (\ie uniform reshape) and losing most visual details (\eg downsampling to 300x300) \citep{xie2019selftraining}. At the same time, samples of the same class have little intra-class variance, while being affected by large contextual changes (background, scale, perspective, illumination, \etc). To contribute in a direction which has not yet been properly addressed by the AI community, in this paper we present a visual challenge which is different in all these aspects. It is based on museum art mediums (MAMe), where attention to detail is essential, where there is huge intra-class variance, and where contextual information is not a factor.

The properties of ImageNet and ImageNet-like datasets have popularized the practice of interpolating images. This approach allows to reduce the memory requirements of models, avoiding high resolution (HR) images, and removing the hindrances of variable-shaped (VS) inputs. The first CNN models tackling the ImageNet challenge interpolated images to a fixed size of 224x224 pixels~\citep{simonyan2014very, szegedy2015going}. More recent solutions increased that size to 229x229~\citep{szegedy2016rethinking}, 331x331~\citep{zoph2018learning}, 480x480~\citep{huang2019gpipe} or even 600x600~\citep{he2017mask, lin2017feature} pixels, as scaling the image resolution is known to result in better performances on some cases~\citep{tan2019efficientnet,ghosh2019reshaping}. Even so, the nature of the ImageNet-like problems minimized these inconveniences, resulting in competitive performances even when using relatively small input sizes~\citep{xie2019selftraining}. Given the prominence of ImageNet, this particularity biased research. Indeed, beyond this ImageNet-like tasks, there are many current and future visual challenges where exploitation of HR and VS  properties are likely to be relevant for improving performance.

Visual challenges in the medical domain are often based on the identification of small-scale visual patterns, requiring both attention to detail and an understanding of large structures. In domains like breast cancer detection, the benefit of exploiting the highest possible image resolution has already been highlighted \citep{geras2017high, lotter2017multi}, motivating the use of HR images. Similarly, image recognition systems used for autonomous driving also benefits from using HR images, as this entails detection at further distances, which have enormous safety implications. Current solutions already use images that are larger than 0.25 MP \citep{chen2017multi, treml2016speeding}.

The motivation for research on VS images derives from the increasing popularity of crowd-sourced datasets, such as Open Images~\citep{OpenImages}. These datasets combine data produced from multiple sources, which saves time and effort, at the expense of obtaining data in different resolutions and shapes (\eg landscape or portrait). In this context, standard training procedures using squared images are forced to interpolate them, hence deforming their image patterns. These image deformations introduce noise within data, potentially decreasing performance.

% TODO: Llistat de aportacions que fa el paper
The main contribution of this paper is the MAMe dataset itself, which is made available for the research community (\S\ref{sec:mame_dataset}). Beyond extensive statistics and expert insights, this work also provides several baselines based on popular architectures: VGG~\citep{simonyan2014very}, ResNet~\citep{he2016deep}, DenseNet~\citep{huang2017densely} and EfficientNet~\citep{tan2019efficientnet} (\S\ref{sec:baselines}). Further experiments (\S\ref{sec:exp_res}) are performed to assess the impact on accuracy when using high resolution, variable shape or both properties in conjunction. One final experiment (\S\ref{sec:exp_res}) highlights whether performance gain comes from increasing the amount of image information or from increasing the models internal representation (as consequence of increasing the input size). This last experiment provide really different results when using the MAMe dataset in contrast to ImageNet \citep{sandler2019nondiscriminative}, highlighting the particularity of the MAMe dataset. Finally, we provide a qualitative analysis of the MAMe through a set of expert analysis and explainability experiments (\S\ref{sec:exp}).

\begin{figure*}[t]
\centering
\includegraphics[width=\textwidth]{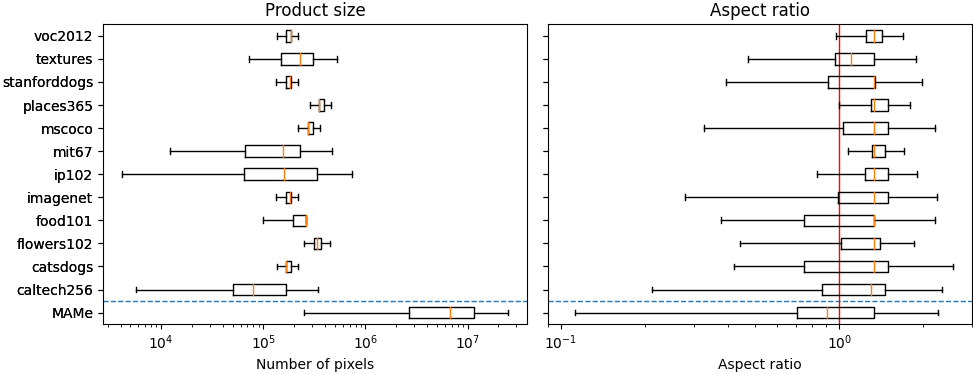}
\caption{Product size and aspect ratio distribution over several datasets, both on log scale. The dashed horizontal blue line separates a sample of current image classification datasets, and the proposed MAMe dataset. The vertical red line at aspect ratio 1.0 shows the border between portrait (left side) and landscape (right side) images.}
\label{fig:ps_ar_distr}
\end{figure*}

\section{Related work}\label{sec:rw}

%-Intro to current datasets.
There are many visual challenge datasets in the current literature. There are however, very few containing images larger than 500x500 pixels, and with a significant variance in their aspect ratio. To illustrate that point we analyze a sample of popular datasets which satisfy three conditions we consider essential for attracting and generating high quality research:

\begin{itemize}
    \item The dataset is publicly available.
    \item The dataset labels are reliable.
    \item The dataset has at least 100 instances per class. 
\end{itemize}

The first requires data to be as public as possible, to reach the largest possible number of researchers. The second one excludes all datasets that contain labels not validated by humans or that have been crowd-labeled, as these may contain a significant amount of noise (and noise reduces the reliability of experimental results). The third enforces a minimum number of instances. We consider this a necessity for thorough research experimentation. We were nonetheless flexible in this regard, as some datasets of those analized contain some classes with less than 100 instances.

The sample analyzed contains the following 12 datasets: ImageNet 2012 \citep{russakovsky2015imagenet}, Food101 \citep{bossard2014food}, IP102 \citep{wu2019ip102}, Places365 \citep{zhou2017places}, Mit67 \citep{quattoni2009recognizing}, Flower102 \citep{nilsback2008automated}, CatsDogs \citep{parkhi2012cats}, StanfordDogs \citep{khosla2011novel}, Textures \citep{cimpoi2014describing}, Caltech256 \citep{griffin2007caltech}, Microsoft COCO \citep{lin2014microsoft} and Pascal VOC 2012 \citep{pascal-voc-2012}. 
For each one we compute the product size (\ie width multiplied by height) and aspect ratio (\ie width divided by height) distributions. For the three datasets with more than 100,000 total samples (ImageNet 2012, Places365 and Microsoft COCO) we use a random sample of 100,000 images. Distributions for all 12 datasets can be seen in Figure \ref{fig:ps_ar_distr}.

In terms of number of pixels (left plot), current image classification datasets rarely contain images with more than 1 megapixel (MP). For reference purposes, none of the 12 datasets contain images bigger than 1,000 x 1,000 pixels, assuming unitary aspect ratio. This already indicates a significant bias in current research, and a mismatch with current technology, as popular image taking resolutions are well above that size. Obviously, there are datasets with images larger than 1 MP, but these are typically either private, unreliability labeled \citep{OpenImages}, or have very few instances per class \citep{GoogleLandmarks}. In this context, as shown at the bottom of Figure \ref{fig:ps_ar_distr}, the MAMe dataset stands out, containing a large volume of reliable labeled HR images. In fact, all images in the Q1-Q3 interval of the MAMe dataset are bigger than the largest image found on all analyzed datasets. The mean image size for the MAMe dataset is 6.6MP (\eg 2350x2350 in a squared image), one order of magnitude larger than all images contained in the analyzed datasets.

Regarding aspect ratio, the right plot of Figure \ref{fig:ps_ar_distr} shows how the majority of images found in current datasets are landscape. All datasets have their median in the landscape side, only half of the datasets contain Q1 within the portrait side, and only 3 contain a significant amount of portrait images (Food101, CatsDogs and Caltech256). However, even these have their aspect ratio distribution clearly skewed towards landscape images (notice that the median is quite close to the third quartile on all three cases). In contrast, the proposed MAMe dataset has a balanced distribution, containing approximately the same number of portrait and landscape images. The aspect ratio distribution is also much wider than the other datasets, showing how the MAMe dataset contains infrequently wide and tall images.

\section{The MAMe dataset}\label{sec:mame_dataset}

In this work we propose the Museum Artworks Medium dataset, abbreviated as the MAMe dataset. MAMe is an image classification dataset focused on the recognition of mediums in artworks and heritage held by museums (\eg \textit{Oil on canvas}, \textit{Bronze} or \textit{Woodcut}). Medium is a broad technical term used to describe several aspects of artworks~\citep{maynor1989paper}. On one hand, it can be used to describe the main physical components used for the creation of an artwork, such as \textit{Oil on canvas}. However, medium can also refer to the technique used to produce the artwork. \textit{Engraving}, for example, is the printed result of engraving a metal plate. Both of these interpretations of medium are freely used by museums to organize their collections.

\begin{figure}[tb]
    \centering
    \begin{tabular}{ccc}
        \includegraphics[width=0.2\textwidth]{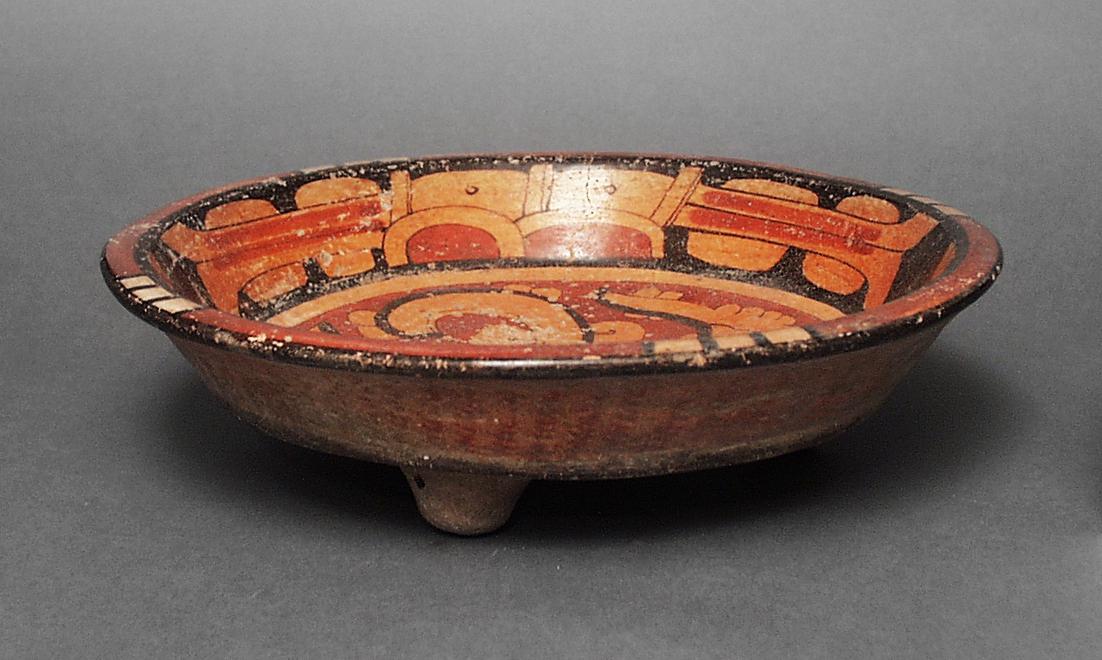} &
        \includegraphics[width=0.2\textwidth]{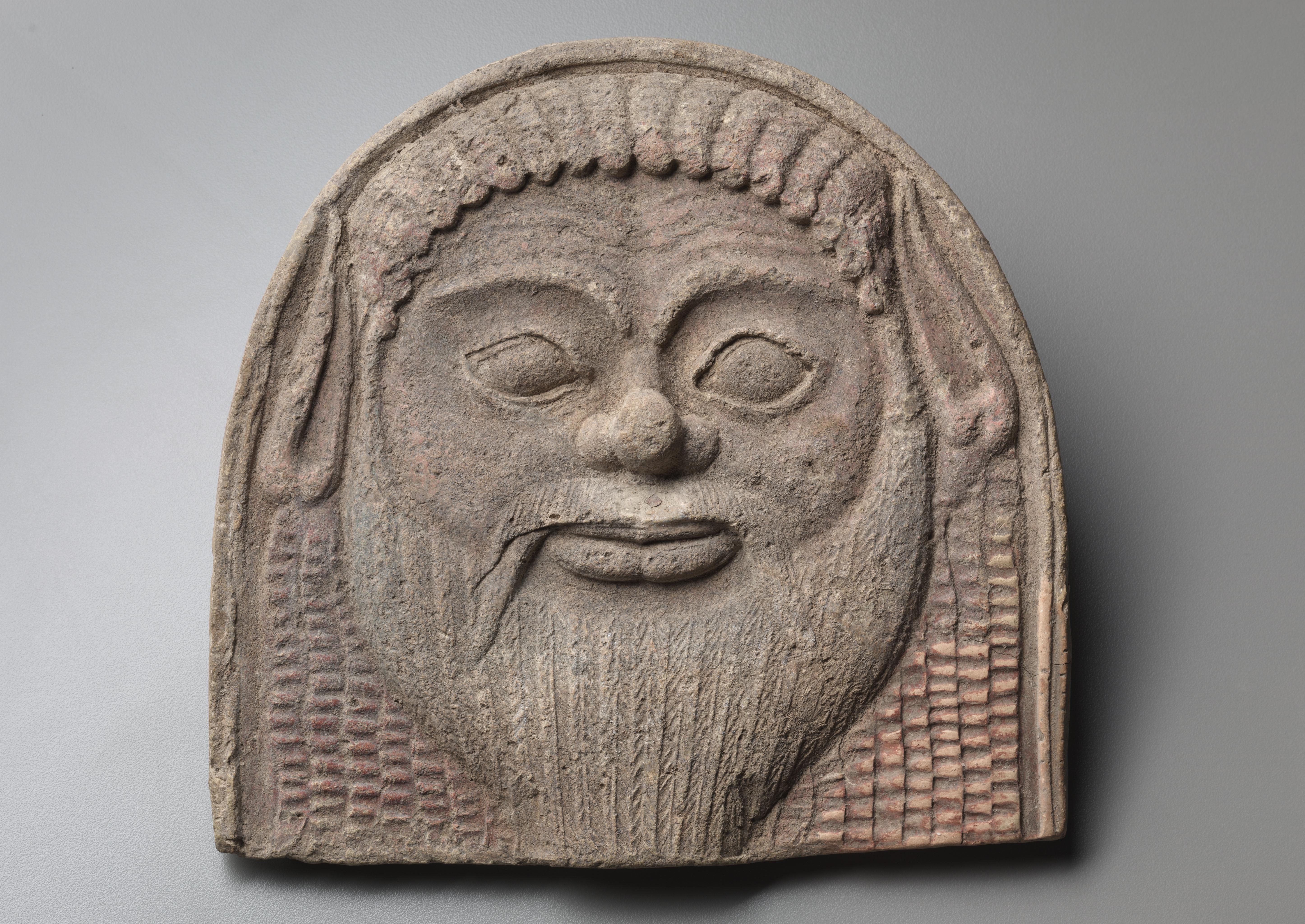} &
        \includegraphics[width=0.2\textwidth]{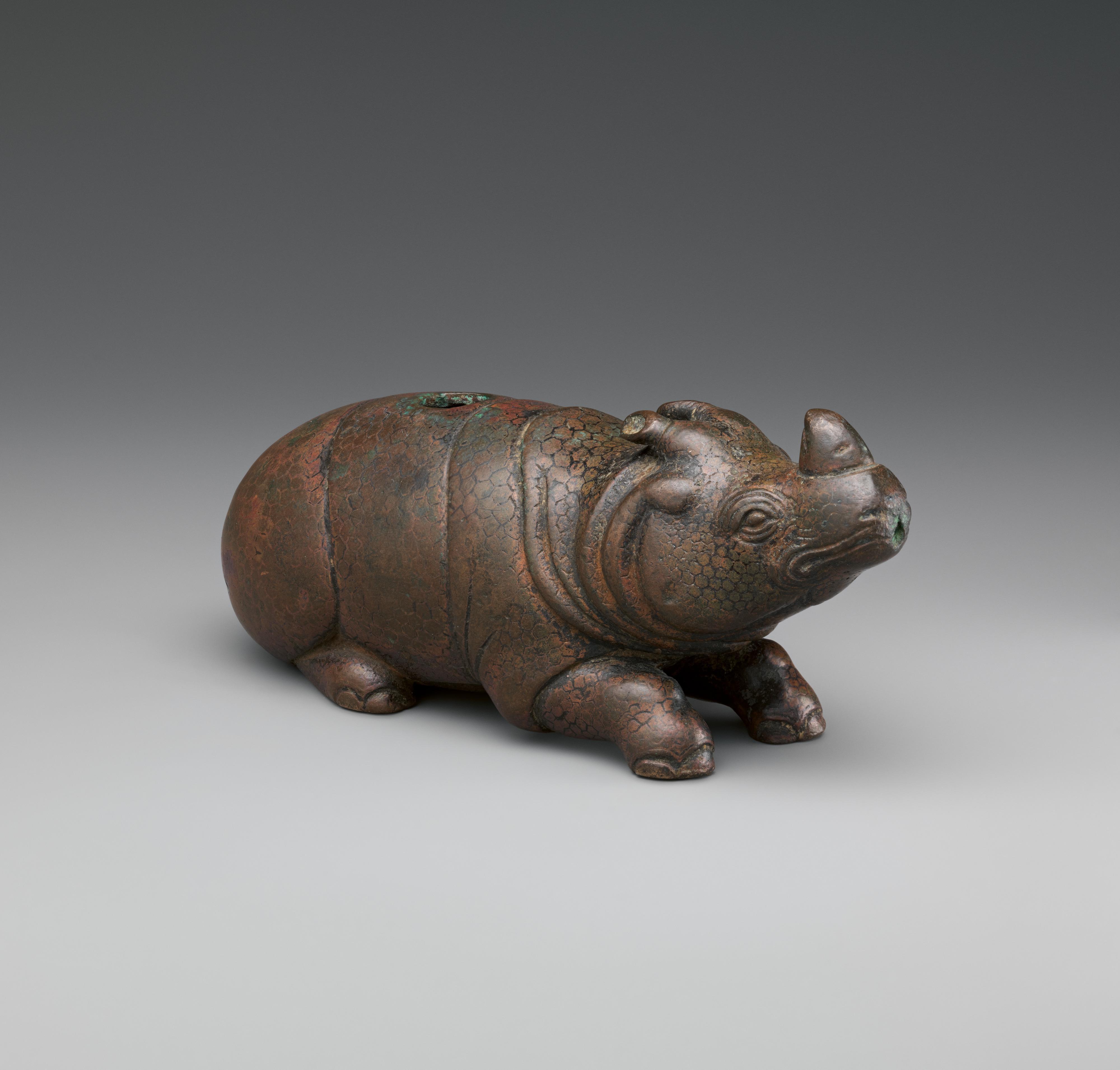} \\
        \includegraphics[height=0.19\textwidth]{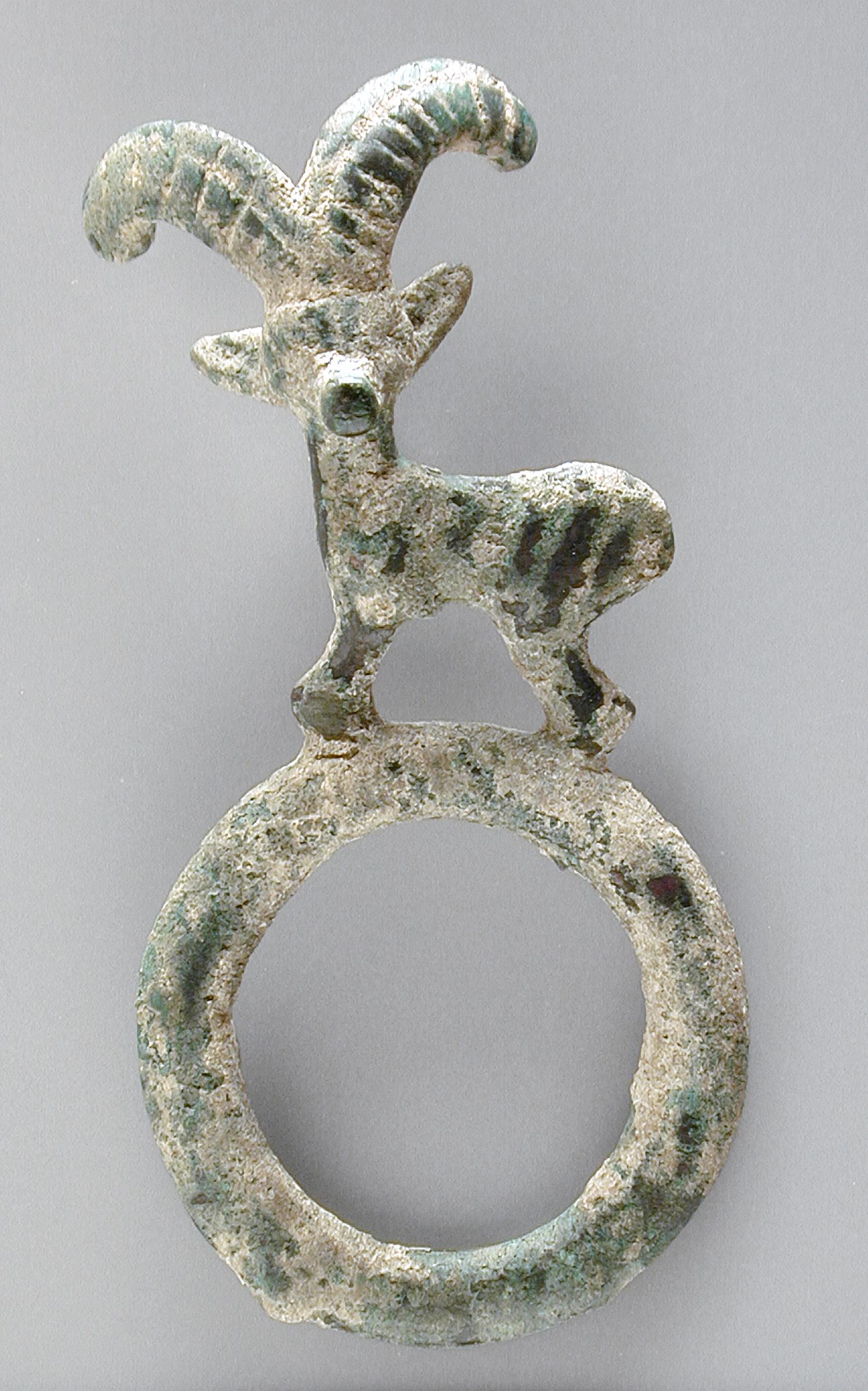} &
        \includegraphics[width=0.2\textwidth]{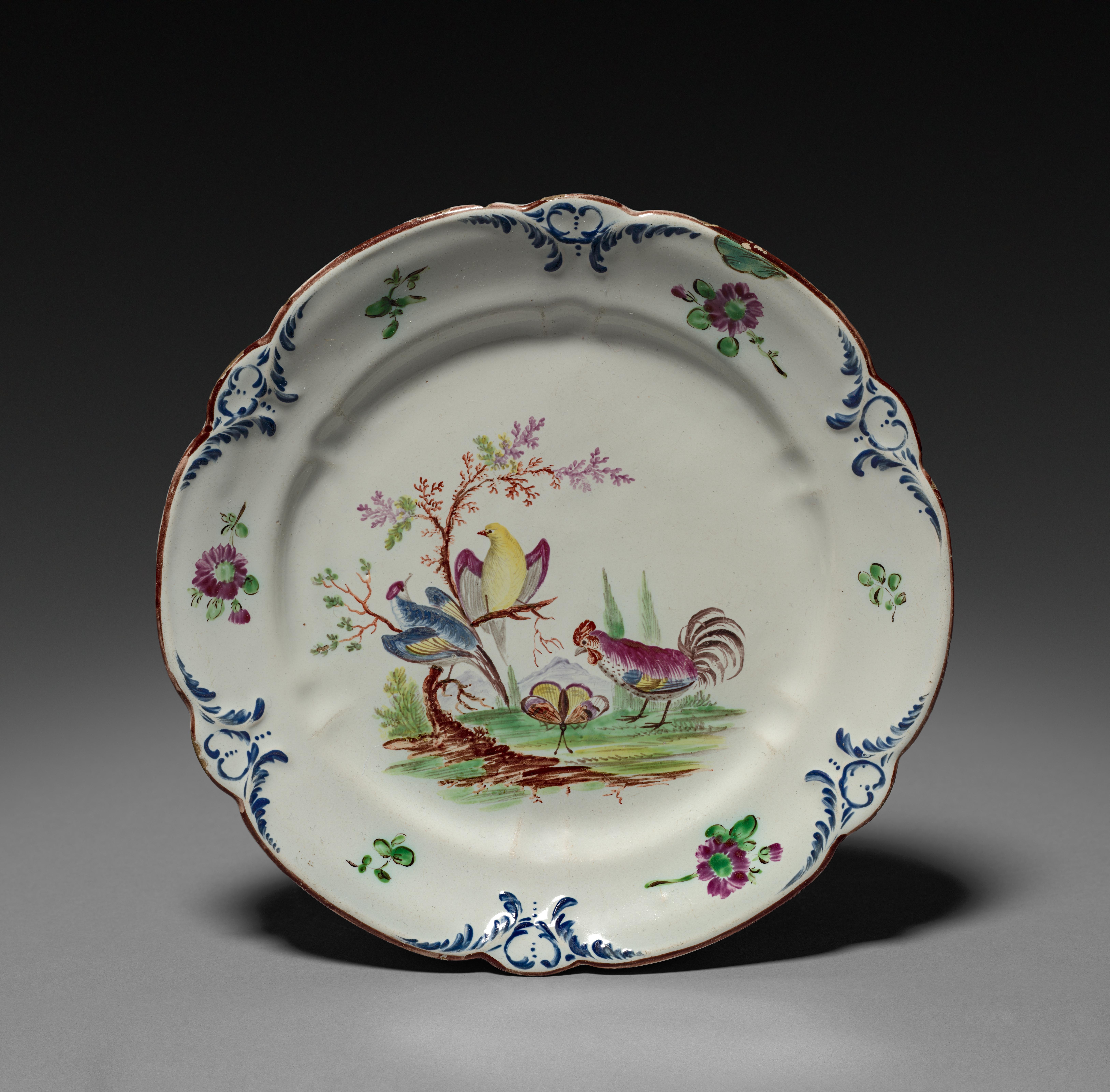} &
        \includegraphics[height=0.19\textwidth]{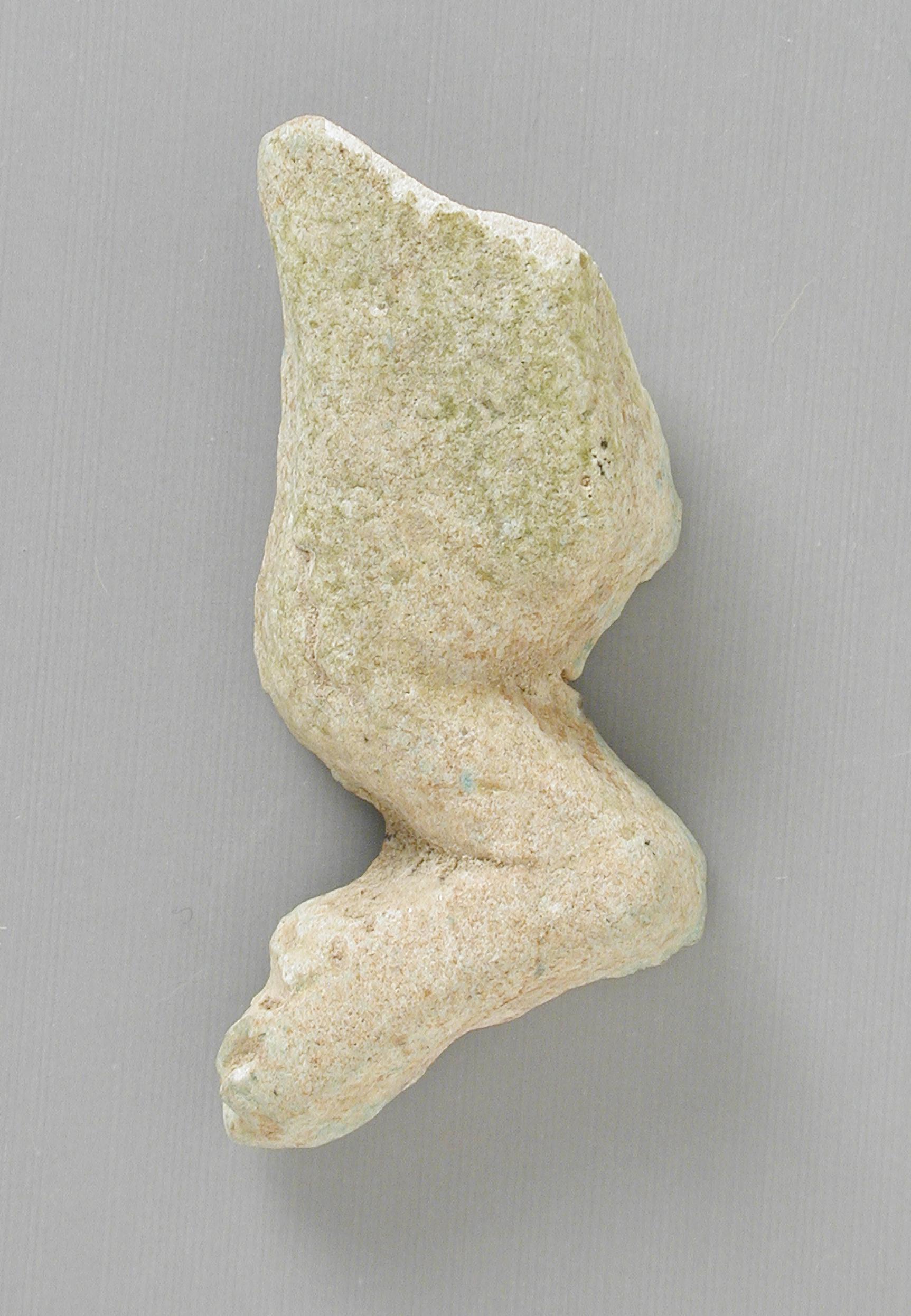}\\
    \end{tabular}
    \caption{Example of intra-class variance. Images in the same row belong to the same medium class, but share few visual features. The first row belongs to \textit{Ceramic}, the second row to \textit{Bronze} and the third row to \textit{Faience}.}
    \label{fig:intra-class_variance}
\end{figure}

As detailed in \S\ref{tab:medium_museums}, the classes considered in the MAMe dataset comprise a wide variety of mediums according to both interpretations of the term. These can range from simple material aspects (\eg \textit{Bronze}, \textit{Silver} or \textit{Gold}) to complex, high-level techniques (\eg \textit{Faience}, \textit{Woodblock} or \textit{Woven fabric}). The variety of relevant features in MAMe requires both attention to detail and to the overall image structure. Meanwhile, the essence of art causes widely different artworks to share the same label. The degree of intra-class variance of MAMe is exemplified in Figure \ref{fig:intra-class_variance}.

\subsection{Data acquisition}

\begin{figure}[tb]
\centering
\includegraphics[width=\textwidth]{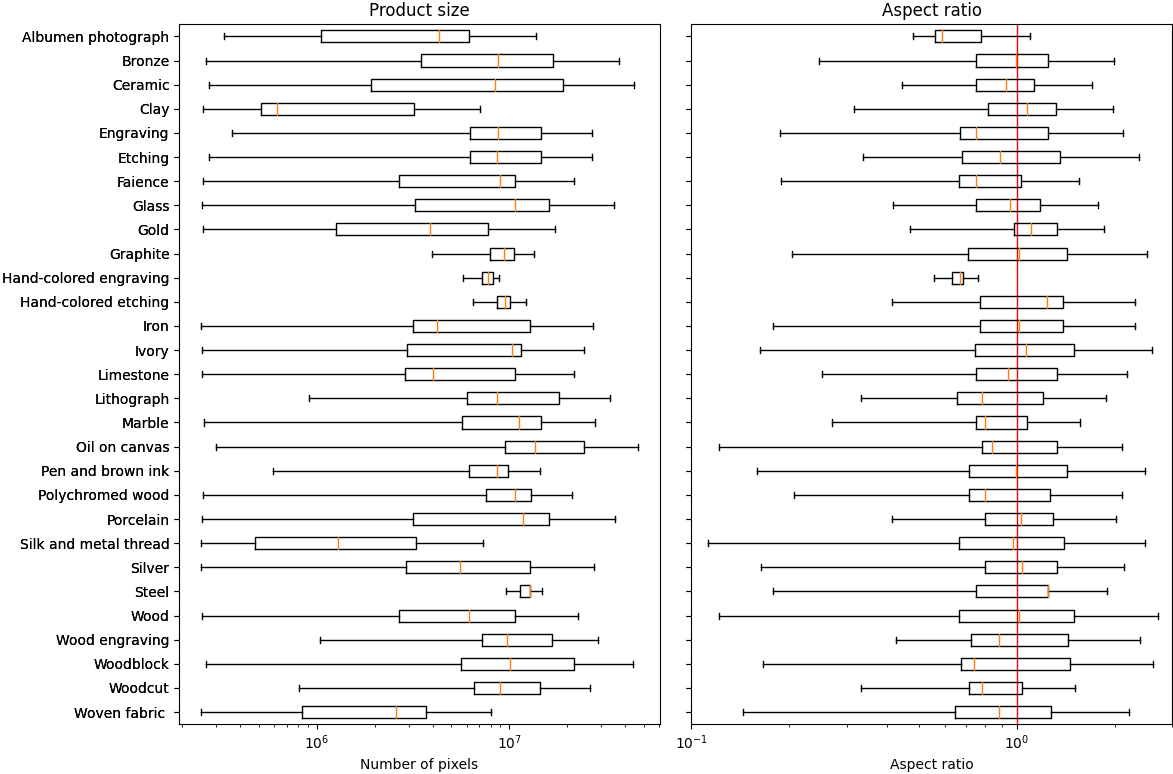}
\caption{Product size and aspect ratio distribution over all classes of the MAMe dataset. Distributions are represented in box-plots, both of them on log scale. The vertical red line at aspect ratio 1.0 shows the border between portrait (left side) and landscape (right side) images.}
\label{fig:class_ps_ar_distr}
\end{figure}

% Data comes from 3 different Museums (LACMA, CLEVELAND and MET). Data released under CC0 license. Common trend between Museums around the world.
In the past few years, museums around the world have been endorsing the policy of publicly releasing images of their heritage. Some of these museums release HR images under a CC0 license, allowing a free and unrestricted use of the data. We base our work on the data released by three museums. These were chosen because all three endorse the CC0 license, include a large number of images, provide accessible labels for them, and make it feasible to access their data in an automatized manner:
\begin{itemize}
    \item The Metropolitan Museum of Art of New York (from now on the Met museum)~\citep{Metrelease}.%: Founded in 1870, it is recognized as one of the most important museums in the world, due to the size of its collections and the importance of the works it holds.
    \item The Los Angeles County Museum of Art (from now on the Lacma museum)~\citep{Lacmarelease}.%: It is the largest art museum in the western United States. Although it does not reach the size of the Met, it has important collections. It was part of the Los Angeles Museum of History, Science and Art, founded in 1910.
    \item The Cleveland Museum of Art (from now on the Cleveland museum)~\citep{Clevelandrelease}.%: Founded in 1913, it is one of the most distinguished comprehensive art museums in the United States, internationally recognized for its substantial holdings. 
\end{itemize}

All three museums hold large artistic collections with a general scope, including artworks from all over the world, from very early cultures to recent ones. For accessing the data, the Cleveland museum publishes an API to automatically download images. Lacma and Met on the other hand provide access to their images only through their webpages. This implies an image-by-image download process, for which we built museum-specific crawlers. By these means we downloaded approximately 232,000 images from the Met museum, 26,000 from the Lacma museum and 32,000 from the Cleveland museum. From this data, we define the MAMe dataset, composed by an expertly-curated subset of the data. The final selection includes 37,407 images belonging to 29 classes. The class selection process was made following several technical criteria, including balance between museums (to avoid potential bias), balance and volume of class instances (to facilitate research), and image resolution (to enable HR exploration). Grey scale images were discarded. Significantly, museum images have a natural tendency towards VS (\eg human sculptures tend to be tall, while paintings tend to be wide). Although we did not encouraged its presence, this natural feature is shown in the dataset statistics (see right plot of Figure~\ref{fig:class_ps_ar_distr}).

\begin{table}[t]
\centering
\caption{For each medium class within MAMe, distributions of instances among museums. The Met, Lacma and Cleveland museums are labeled as "Met", "Lac" and "Cle" respectively. Museum distributions are divided by data splits, into training, validation and test ("Train", "Val" and "Test" respectively). The last four columns show values aggregated for all data splits ("All"). The "Test" and "All" sections contain a 4th column indicating the total ("Total"). These values are not provided for "Train" and "Val" since these are constant (700 and 50 respectively).}
\label{tab:medium_museums}
\begin{tabular}{l|rrr|rrr|rrrr||rrrr}
\multicolumn{1}{c|}{Medium} & \multicolumn{3}{c|}{Train} & \multicolumn{3}{c|}{Val} & \multicolumn{4}{c||}{Test} & \multicolumn{4}{c}{All} \\
                           & Met     & Lac    & Cle    & Met    & Lac    & Cle   & Met  & Lac  & Cle  & Total & Met  & Lac & Cle & Total  \\ \hline
Albumen photo & 700 & 0 & 0 & 50 & 0 & 0 & 700 & 0 & 0 & 700 & 1450 & 0 & 0 & 1450 \\
Bronze & 234 & 233 & 233 & 16 & 17 & 17 & 233 & 233 & 234 & 700 & 483 & 483 & 484 & 1450 \\
Ceramic & 242 & 242 & 216 & 17 & 18 & 15 & 241 & 241 & 218 & 700 & 500 & 501 & 449 & 1450 \\
Clay & 695 & 5 & 0 & 49 & 1 & 0 & 310 & 2 & 1 & 313 & 1054 & 8 & 1 & 1063 \\
Engraving & 234 & 233 & 233 & 16 & 17 & 17 & 233 & 234 & 233 & 700 & 483 & 484 & 483 & 1450 \\
Etching & 234 & 233 & 233 & 16 & 17 & 17 & 233 & 234 & 233 & 700 & 483 & 484 & 483 & 1450 \\
Faience & 599 & 63 & 38 & 43 & 5 & 2 & 598 & 63 & 39 & 700 & 1240 & 131 & 79 & 1450 \\
Glass & 576 & 53 & 71 & 41 & 3 & 6 & 575 & 55 & 70 & 700 & 1192 & 111 & 147 & 1450 \\
Gold & 448 & 95 & 157 & 32 & 7 & 11 & 448 & 96 & 156 & 700 & 928 & 198 & 324 & 1450 \\
Graphite & 565 & 8 & 127 & 40 & 0 & 10 & 151 & 3 & 34 & 188 & 756 & 11 & 171 & 938 \\
H-C engraving & 30 & 641 & 29 & 3 & 45 & 2 & 14 & 300 & 14 & 328 & 47 & 986 & 45 & 1078 \\
H-C etching & 699 & 1 & 0 & 50 & 0 & 0 & 582 & 2 & 0 & 584 & 1331 & 3 & 0 & 1334 \\
Iron & 569 & 2 & 129 & 40 & 0 & 10 & 215 & 1 & 49 & 265 & 824 & 3 & 188 & 1015 \\
Ivory & 611 & 31 & 58 & 43 & 2 & 5 & 498 & 27 & 47 & 572 & 1152 & 60 & 110 & 1322 \\
Limestone & 593 & 56 & 51 & 42 & 5 & 3 & 591 & 56 & 53 & 700 & 1226 & 117 & 107 & 1450 \\
Lithograph & 277 & 147 & 276 & 19 & 11 & 20 & 276 & 148 & 276 & 700 & 572 & 306 & 572 & 1450 \\
Marble & 520 & 86 & 94 & 37 & 6 & 7 & 190 & 32 & 35 & 257 & 747 & 124 & 136 & 1007 \\
Oil on canvas & 265 & 171 & 264 & 18 & 12 & 20 & 264 & 172 & 264 & 700 & 547 & 355 & 548 & 1450 \\
P\&B ink & 665 & 12 & 23 & 47 & 1 & 2 & 271 & 6 & 9 & 286 & 983 & 19 & 34 & 1036 \\
Poly wood & 525 & 59 & 116 & 37 & 4 & 9 & 281 & 32 & 62 & 375 & 843 & 95 & 187 & 1125 \\
Porcelain & 447 & 56 & 197 & 31 & 4 & 15 & 446 & 57 & 197 & 700 & 924 & 117 & 409 & 1450 \\
S\&M thread & 680 & 0 & 20 & 48 & 0 & 2 & 92 & 1 & 2 & 95 & 820 & 1 & 24 & 845 \\
Silver & 452 & 81 & 167 & 32 & 5 & 13 & 450 & 83 & 167 & 700 & 934 & 169 & 347 & 1450 \\
Steel & 628 & 0 & 72 & 44 & 0 & 6 & 118 & 1 & 14 & 133 & 790 & 1 & 92 & 883 \\
Wood & 577 & 43 & 80 & 41 & 3 & 6 & 576 & 44 & 80 & 700 & 1194 & 90 & 166 & 1450 \\
Wood engraving & 410 & 15 & 275 & 29 & 1 & 20 & 211 & 9 & 141 & 361 & 650 & 25 & 436 & 1111 \\
Woodblock & 259 & 258 & 183 & 18 & 19 & 13 & 258 & 258 & 184 & 700 & 535 & 535 & 380 & 1450 \\
Woodcut & 417 & 51 & 232 & 30 & 3 & 17 & 416 & 52 & 232 & 700 & 863 & 106 & 481 & 1450 \\
Woven fabric & 658 & 3 & 39 & 46 & 0 & 4 & 656 & 5 & 39 & 700 & 1360 & 8 & 82 & 1450
\end{tabular}
\end{table}

\subsection{Label mapping}

All three museums (Met, Lacma and Cleveland) reported the medium used to represent their artworks as metadata. Unfortunately, there is not a unique ontology behind, as each museum uses a different level of detail and interpretation of medium. Some mediums are subtypes of another mediums. Some mediums are reported under different names. And some mediums are combinations of other mediums. Experts from the art domain grouped the medium metadata into coherent classes, following their professional understanding of artistic coherency and visual discriminability. Classes which could not be discriminated visually by a human without technical aid (\eg a microscope) were discarded. The main expert criteria used to determine the classes are the following:

\begin{itemize}
    \item Written coherency: Medium categories written in different forms refering to the same term are aggregated (\eg \textit{Bronze} and \textit{bronze})
    \item Terminology coherency: Medium categories which are considered to be analogous are aggregated (\eg \textit{Ceramic} and \textit{Pottery}).
    \item Taxonomic coherency: Object belonging to the same parent medium are sometimes aggregated (\eg \textit{Terracotta} and \textit{Ceramic}). Where technical criteria allows, medium subtypes are left as a separate class (\eg \textit{Porcelain}).
    \item Visual coherency: Medium categories which cannot be visually differentiated at plain sight are aggregated (\eg \textit{Hard-paste porcelain} and \textit{Soft-paste porcelain} into \textit{Porcelain}, \textit{Cotton} and \textit{Linen} into \textit{Woven fabric}).
\end{itemize} 

After enforcing a minimum amount of 850 samples per medium (adding up train, val and test), the MAMe dataset contains 29 different classes. These are shown in the left column of Table \ref{tab:medium_museums}. Notice we made an exception with the \textit{Silk and metal thread} medium, which only contains 845 samples. A detailed description of the nature of each class is provided in Table \ref{tab:medium_descriptions}. Visual details on how to discriminate some of these classes are discussed in \S\ref{sec:exp}.

\begin{table}[t]
    \centering
    \caption{Descriptions of the medium classes. Some descriptions are obtained from the museum sources~\citep{Metdescriptions}.}
    \label{tab:medium_descriptions}
    \begin{adjustbox}{width=\textwidth}
        \begin{tabular}{l|l}
\textit{Medium} & \textit{Description} \\ \hline
\begin{tabular}[c]{@{}l@{}}Albumen\\ photograph\end{tabular} & \begin{tabular}[c]{@{}l@{}}Photographic prints on paper support. Paper is coated with egg white and\\ silver nitrate, and exposed to sunlight in contact with a glass negative.\end{tabular} \\ \hline
Bronze & 
\begin{tabular}[c]{@{}l@{}}Objects mainly made of bronze (cooper and tin alloy).\\ Includes both polished and hammered bronze.\end{tabular} \\ \hline
Ceramic & 
\begin{tabular}[c]{@{}l@{}}Includes pottery, stoneware, earthware and terracotta.\\ It may include glazed, slip-painted or painted textures.\end{tabular} \\ \hline
Clay & 
\begin{tabular}[c]{@{}l@{}}Objects made of clay or mud.\\ In most cases the object has not been baked, or it has at very low temperatures.\end{tabular} \\ \hline
Engraving & \begin{tabular}[c]{@{}l@{}}Intaglio printmaking process in which lines are cut into a metal plate in order to hold the ink. The plate\\ can be made of copper or zinc. \end{tabular} \\ \hline
Etching & \begin{tabular}[c]{@{}l@{}}Intaglio printmaking process in which lines or areas are incised using acid into a metal plate in order\\ to hold the ink. The plate can be made of iron, copper, or zinc.\end{tabular} \\ \hline
Faience & May contain egyptian faience (sintered quartz with a vitreous coating) or tin-glazed pottery.   \\ \hline
Glass & Objects mainly made of glass (eg blown, or pressed). Stained glass windows are excluded. \\ \hline
Gold & Objects mainly made of gold. Includes polished gold, hammered gold and other surface textures. \\ \hline
Graphite & Drawings or sketches made with graphite lead on paper. \\ \hline
\begin{tabular}[c]{@{}l@{}}Hand-colored\\ engraving\end{tabular} & \begin{tabular}[c]{@{}l@{}}Engraving prints hand-colored after the printmaking process. Prints are colored using either\\ watercolor or wash techniques.\end{tabular} \\ \hline
\begin{tabular}[c]{@{}l@{}}Hand-colored\\ etching\end{tabular} & \begin{tabular}[c]{@{}l@{}}Etching prints hand-colored after the printmaking process. Prints are colored using either\\ watercolor or wash techniques.\end{tabular} \\ \hline
Iron & Objects mainly made of iron. Includes polished iron, hammered iron and other surface textures. \\ \hline
Ivory & \begin{tabular}[c]{@{}l@{}}Objects made mainly of ivory (elephant or walrus tusks).\\ Includes watercolor on ivory miniature portraits (medallions).\end{tabular} \\ \hline
Limestone & Objects mainly made of limestone, a sedimentary rock mainly composed by calcium carbonate. \\ \hline
Lithograph & \begin{tabular}[c]{@{}l@{}l@{}}Planographic printmaking process in which a design is drawn onto a flat stone (or prepared\\ metal plate, usually zinc or aluminum) and affixed by means of a chemical reaction. May contain\\ lithographic offset prints and hand-colored monochrome lithographs. \end{tabular} \\ \hline
Marble & Objects mainly made of marble, a metamorphic rock composed of calcite or dolomite. \\ \hline
Oil on canvas & \begin{tabular}[c]{@{}l@{}}Fabric streched into frame (stretcher bar), with a preparation layer (or ground layer) painted\\ with linseed oil and pigment.\end{tabular} \\ \hline
\begin{tabular}[c]{@{}l@{}}Pen and\\ brown ink\end{tabular} & \begin{tabular}[c]{@{}l@{}}Drawings or sketches on paper, mainly made in brown ink (either with a dip pen, a fountain\\ pen or a brush). Can be supplemented by other procedures such as wash (brown or black ink)\\ or dry media. Some artworks may contain aged iron gall ink, or other similar\\ brown inks such as bister or sepia ink.\end{tabular} \\ \hline
\begin{tabular}[c]{@{}l@{}}Polychromed\\ wood\end{tabular} & \begin{tabular}[c]{@{}l@{}}Objects made of painted wood. Includes three-dimensional objects and painted surfaces,\\ such as panel painting (oil on wood or tempera on wood).\end{tabular} \\ \hline
Porcelain & \begin{tabular}[c]{@{}l@{}}A type of ceramic composed by quartz, feldspar and kaoli cooked at high temperatures.\\ May contain soft-past porcelain.\end{tabular}  \\ \hline
\begin{tabular}[c]{@{}l@{}}Silk and\\ metal thread\end{tabular} & Woven fabric objects made of silk with metallic threads, typically forming an embroidery. \\ \hline
Silver & Objects mainly made of silver. Includes both polished and hammered silver. \\ \hline
Steel & Objects mainly made of steel (alloy of iron with carbon). \\ \hline
Wood & Non polychromed wood objects. Inlcudes several wood types such as oak, boxwood or limewood. \\ \hline
\begin{tabular}[c]{@{}l@{}}Wood\\ engraving\end{tabular} & A type of woodcut printmaking process characteristic for using a block cut along the end-grain. \\ \hline
Woodblock & \begin{tabular}[c]{@{}l@{}}A type of woodcut printmaking process typically used by oriental cultures. This type of woodcut\\ is carved along the wood grain and uses a different block for each color printed.\end{tabular} \\ \hline
Woodcut & \begin{tabular}[c]{@{}l@{}}The oldest form of printmaking. Relief process in which knives and other tools are used to\\ carve a design into the surface of a wooden block. The raised areas that remain after the block\\ has been cut are inked and printed, while the recessed areas that are cut away\\ do not retain ink, and will remain blank in the final print.\end{tabular} \\ \hline
Woven fabric & \begin{tabular}[c]{@{}l@{}}Fabric objects woven with a loom. Includes linen, cotton, silk and others.\\ Fabrics appear in several forms such as plain fabrics, embroideries or printed fabrics.\end{tabular}
\end{tabular}
    \end{adjustbox}
\end{table}

\subsection{Dataset details}

% The dataset is publicly available in the follwoing website
The MAMe dataset is publicly available~\footnote{\label{note1}\url{https://hpai.bsc.es/MAMe-dataset}}. The site provides access to all the original images, and a CSV file with metadata for each of them. This metadata includes the following information:
\begin{itemize}
    \item the \textbf{image filename}
    \item the \textbf{medium} of the artwork (\ie the classification label)
    \item the \textbf{museum} from where the image was obtained
    \item the artwork \textbf{ID} given by the museum
    \item the \textbf{data split} of the instance (\ie train, validation or test set)
    \item the \textbf{width} of the image
    \item the \textbf{height} of the image
    \item the \textbf{product size} of the image (\ie width multiplied by height)
    \item the \textbf{aspect ratio} of the image (\ie width divided by height)
\end{itemize}

The dataset contains 29 medium classes. Each class is composed by at least 850 images and, at most 1,450. Each class contains 700 images for training, 50 images for validation and a variable amount of images for the test set (\ie the test set is unbalanced). The minimum amount of instances in the test set is 100 (except for \textit{Silk and metal thread} with 95) and the maximum is 700.
In total, the MAMe dataset is composed by 37,407 HR images. All images in the MAMe dataset have, at least, a resolution of 0.25MP, equivalent to a squared image of 500x500 pixels. The mean resolution is around 10.3MP, corresponding to an image of more than 3,200x3,200 pixels, and the greatest image has more than 370MP corresponding to an image of 32,683x11,412 pixels (check Figure \ref{fig:ps_ar_distr}). The 37,407 images are divided in subsets as follows: 20,300 images for training and 1,450 images for validation and 15,657 for test. Of those, 24,911 images originate from the Met museum, 5,531 images from the Lacma museum and 6,965 images from the Cleveland museum. An effort was made to keep the data coming from the different museums as balanced as possible, to minimize the possibility of potential biases generated by the nature of artworks and the image taking particularities of each museum. The exact distributions of images per museum, class and data split are shown in Table \ref{tab:medium_museums}. To assess the internal balance of MAMe with regards to HR and VS features, Figure \ref{fig:class_ps_ar_distr} shows the product size and aspect ratio distributions for each medium class. Besides a few classes with particularly narrow or skewed distributions, most of the categories include a wide variety of product sizes and aspect ratios.

\section{Baselines and Experiments}\label{sec:baselines}

This section introduces and evaluates both a set of baseline models and a set of hypothesis. The purpose of baselines is to illustrate how the task proposed is coherently constructed (\ie solvable) and worth receiving the attention of researchers. To this end, we employ prototypical solutions from the literature that provide good results on other challenges, and report their performance on the MAMe dataset. 
%Baselines also offer a reference to future contributions, a performance to compare against and to measure progress.

\begin{figure}[t]
    \centering
    \includegraphics[width=\textwidth]{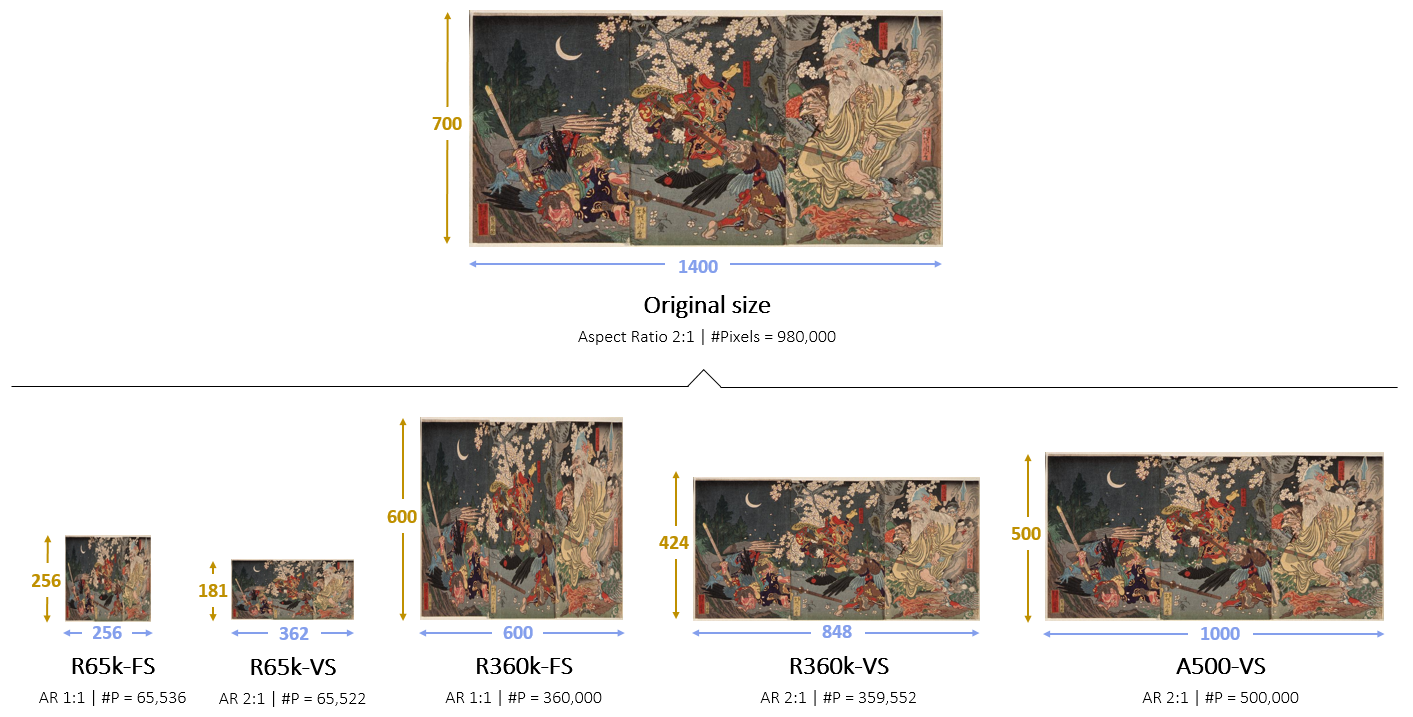}
    \caption{Visualization of the different MAMe data types, exemplified for one particular instance. FS stands for fixed shape and VS for variable shape. R stands for resolution and A for axis.}
    \label{fig:data_type}
\end{figure}

Additionally, to highlight differences of high resolution (HR) and variable shape (VS) properties \wrt low resolution (LR) and fixed shape (FS) in the context of MAMe, we perform a set of experiments. These are designed to evaluate the following hypothesis:

\begin{hyp}\label{hyp:first}
MAMe benefits from HR data \wrt LR data.
\end{hyp}

\begin{hyp} \label{hyp:second}
MAMe benefits from VS data \wrt FS data.
\end{hyp}

\begin{hyp} \label{hyp:third}
MAMe benefits from information gain \wrt only resolution gain.
\end{hyp}

All baseline models, hypothesis experiments and the code needed to replicate results are publicly available \footnote{\url{https://github.com/HPAI-BSC/MAMe-baselines}}.

\subsection{MAMe data types}\label{sec:data_type}

Most of current solutions in the literature use squared images to feed their models, that is images with a fixed shape. Additionally, these squared images are typically of low resolution. Resolutions used are diverse, but the most common is 256x256 pixels, corresponding to a total amount of 65,536 pixels. For referencing purposes, we use this data type as a starting point and we call it R65k-FS. For comparison purposes, we use a second data type using the same resolution (\ie same amount of total pixels) but keeping the original aspect ratio of the image, that is with the VS property. This second data type is called R65k-VS. We also produce the HR versions of these two data type. These HR versions contain a total of 360,000 pixels. They are the R360k-FS and the R360k-VS version. Notice that R360k-FS corresponds exactly to an squared image of 600x600 pixels, while R360k-VS contains images of variable shape but fixed number of pixels. See Figure \ref{fig:data_type} for an illustration of all data types used.

The final list of data types used in this section is as follows:
\begin{itemize}
    \item R65k-FS: images are downsampled to 256 x 256 pixels, corresponding to an image size of 65,536 pixels.
    \item R65k-VS: the original aspect ratio is maintained forcing the total number of pixels to 65,536. 
    \item R360k-FS: images are resized to 600 x 600 pixels (360,000 pixels)
    \item R360k-VS: images are rescaled to a total number of pixels to 360,000, maintaining the original aspect ratio.
\end{itemize}

\subsection{Training configurations}\label{sec:base_arc}
% Presenting architectures
In this work we use very well-known architectures: VGG~\citep{simonyan2014very}, ResNet~\citep{he2016deep}, DenseNet~\citep{huang2017densely} and EfficientNet~\citep{tan2019efficientnet}. The specific architecture versions that we use are the following:
% For the baselines of section \ref{sec:baselines_perf} and experiments of section \ref{sec:exp_res}, we use the following versions from these architectures:
\begin{itemize}
    \item VGG11 (configuration A)
    \item VGG16 (configuration D)
    \item ResNet18
    \item ResNet50
    \item EfficientNet-B0
    \item EfficientNet-B3
    \item DenseNet121
\end{itemize}

% For the experiments of section \ref{sec:exp_res} we only use the VGG11, ResNet18, EfficientNet-B0 and EfficientNet-B3 architectures. This is due to the high memory requirements of some of the data types used (\ie R360k-FS, R360k-VS).
% \fnote{Si volem deixar DenseNet121 a R65k, potser hauriem de organitzar aquest tros dient que }

% For experiments, the MAMe data is used in various forms (check subsection \ref{sec:data_type}, some of them requiring high amounts of memory (\eg R360k-FS, R360k-VS, A500-VS). In those extreme cases, we are forced to use shallow versions from these architecture designs, specifically VGG11 (configuration A) and ResNet18 architectures. For DenseNet architecture we still use its the DenseNet121 version.

% Data processing
Our baselines and experiments use several types of input processing. This is divided into two main components: data processing and data augmentation. The first refers to all image transformations required to obtain each data type (according to subsection \ref{sec:data_type}), while the second provides regularization during the training process. The data augmentation is independent of the data type, and is defined as follows:
\begin{enumerate}
    \item Random rotation of the image from [-30, 30] degrees.
    \item Random crop of (0.875 x width, 0.875 x height) pixels. Width and height refer to current dimensions at this point of the processing.
    \item Random horizontal flip with 50\% chance.
\end{enumerate}

As a final step, images are normalized to have values in the range [0, 1] and standardized with $\mu = 0.5$ and $\sigma = 0.5$ (same value for all three channels). Notice that during validation, the data augmentation is adapted. In this phase, steps 1 and 3 are avoided and, step 2 does a center crop instead of random one.

We use the AMSGrad optimizer~\citep{reddi2019convergence} for all the baselines and experiments, a variant of the original Adam optimizer~\citep{kingma2014adam}. Batch sizes and learning rates are optimized for each training, considering memory limitations, training speed and learning convergence. Executions are conducted in a single computing node of the CTE-Power9 cluster at the Barcelona Supercomputing Center, with the following characteristics:
\begin{itemize}
    \item 2 Sockets x IBM Power9 8335-GTH @ 2.4GHz (20 cores and 4 threads/core, total 160 threads).
    \item 4 x GPU NVIDIA V100 (Volta) with 16GB HBM2.
\end{itemize}

\subsection{Baselines performance} \label{sec:baselines_perf}

To show the feasibility and evaluate the difficulty of the MAMe task, we introduce a set of baseline models. Their purpose is to reach the best possible performance using current prototypical solutions. To that end, we employ the following CNN architectures: VGG11, VGG16, ResNet18, ResNet50, EfficientNet-B0, EfficientNet-B3 and DenseNet121. We train these using the four MAMe data types explained in subsection \ref{sec:base_arc}. Due to memory limitations, we only use a subset of architectures on the R360k data type: VGG11, ResNet18, EfficientNet-B0 and EfficientNet-B3. All baselines are trained on top of the corresponding ImageNet pre-trained models \footnote{\url{https://github.com/pytorch/vision}}\footnote{\url{https://github.com/lukemelas/EfficientNet-PyTorch}}. Top 10 baseline results are shown in Table \ref{tab:baseline_results}. These reported results correspond to the mean per class test accuracy using the models achieving minimum validation loss.

\begin{table}[t]
    \begin{minipage}{.49\linewidth}
        \centering
        \caption{Top 10 baseline results for the MAMe dataset. Notice the prevalence of high resolution to a great extent.}
        \label{tab:baseline_results}
% \usepackage{booktabs}

\begin{tabular}{@{}llll@{}}
\multicolumn{4}{c}{}                                                        \\ %\midrule
\textbf{Architecture} & \textbf{Resolution} & \textbf{Shape} & \textbf{Accuracy}   \\ \midrule
EfficientNet-B3       & R360k               & FS             & 88.95\%             \\  \midrule
EfficientNet-B0       & R360k               & FS             & 88.25\%             \\ \midrule
Resnet18              & R360k               & FS             & 88.15\%             \\ \midrule
VGG11                 & R360k               & VS             & 85.42\%             \\ \midrule
EfficientNet-B3       & R65k                & FS             & 85.11\%             \\ \midrule
VGG11                 & R360k               & FS             & 85.04\%             \\ \midrule
Resnet18              & R360k               & VS             & 84.59\%             \\ \midrule
Resnet50              & R65k                & FS             & 84.29\%             \\ \midrule
Resnet50              & R65k                & VS             & 84.07\%             \\ \midrule
EfficientNet-B0       & R65k                & FS             & 83.73\%             \\ \midrule
\end{tabular}

    \end{minipage}
    \begin{minipage}{.49\linewidth}
        \centering
        \caption{Experiment results for \hypref{hyp:first} (more resolution is better) and \hypref{hyp:second} (less deformation is better) hypotheses. \hypref{hyp:first} is assessed vertically (same shape policy, variable resolution), while \hypref{hyp:second} is assessed horizontally (same resolution, variable shape policy).}
        \label{tab:hyp1_2_results}
        \begin{tabular}{@{}r|l|l|l@{}}

                       & \multicolumn{1}{c}{\textbf{FS}} & \textbf{VS} & \multicolumn{1}{c}{\textbf{Architecture}}  \\ \midrule
\multirow{7}{*}{\textbf{R65k}}  & 81.35\%                     & 81.39\%          & VGG11                            \\
                                & 81.20\%                     & 81.21\%          & VGG16                            \\
                                & 83.33\%                     & 82.66\%          & ResNet18                         \\
                                & 84.29\%                     & 84.07\%          & ResNet50                         \\
                                & 73.14\%                     & 76.06\%          & DenseNet121                      \\
                                & 83.73\%                     & 82.38\%          & EfficientNet-B0                  \\
                                & 85.11\%                     & 83.48\%          & EfficientNet-B3                  \\ \midrule
\multirow{4}{*}{\textbf{R360k}} & 85.04\%                     & 85.42\%          & VGG11                            \\
                                & 88.15\%                     & 84.59\%          & ResNet18                         \\
                                & 88.25\%                     & 79.11\%          & EfficientNet-B0                  \\
                                & 88.95\%                     & 80.04\%          & EfficientNet-B3                                      
\end{tabular}

    \end{minipage}
\end{table}

After training multiple models with combinations of seven architectures and four MAMe data types, finetuning each of the training models to optimize performance and using pre-trained models from ImageNet, most models achieve accuracies above 80\%. The maximum 88.95\% accuracy is obtained by the EfficientNet-B3 architecture on the R360k-FS data type. These results show that, indeed, the MAMe task is solvable. It also clearly illustrates the benefits of using higher resolutions, as the top 4 models are based on R360k data types. On the other hand, it seems that models are not properly exploiting the VS property, as only VGG11 manages to be in the top 4 with that shape policy. Next, let us assess the relevance of these properties in further detail.

\subsection{Hypothesis evaluation} \label{sec:exp_res}

% 5.2
In this section we aim to validate the three hypothesis introduced in section \ref{sec:baselines}. Our first hypothesis is \textit{\hypref{hyp:first}: MAMe benefits from HR data \wrt LR data}. Since HR contains extra information that is not present in LR data, this hypothesis aims to measure to which degree is this additional information relevant for improving performance on MAMe.

To test \hypref{hyp:first} we train a set of models using R65k-FS and R360k-FS MAMe data types, where the only difference is the resolution of images. Notice both data types share the same proportional distortion \wrt the original shape of images. We do the same experiment using variable shape (\ie R65k-VS and R360k-VS data types). In this case the only difference is the resolution of images because there is no distortion added to the aspect ratio. The architectures used for validating this hypothesis are VGG11, ResNet18, EfficientNet-B0 and EfficientNet-B3. We use this subset of shallow architectures due to the high-memory requirements when using R360k data. The models are trained starting from their corresponding ImageNet pre-trained models. Results are shown in Table \ref{tab:hyp1_2_results} and Table \ref{tab:hyp1_diff}.

\begin{table}[t]
    \begin{minipage}{.49\linewidth}
        \centering
        \caption{Difference in performance between models trained using R65k and R360k data. This results are used to validate hypothesis \hypref{hyp:first}.}
        \label{tab:hyp1_diff}
        \begin{tabular}{@{}r|l|l|l@{}}

                       & \multicolumn{1}{c}{\textbf{FS}} & \textbf{VS} & \multicolumn{1}{c}{\textbf{Architecture}}  \\ \midrule
\multirow{4}{*}{\begin{tabular}[c]{@{}c@{}}\textbf{R65k}\\ \textbf{to} \\ \textbf{R360k}\end{tabular}} & +3.69\%                     & +4.03\%          & VGG11                            \\
                                & +4.82\%                     & +1.93\%          & ResNet18                         \\
                                & +4.52\%                     & -3.27\%          & EfficientNet-B0                  \\
                                & +3.84\%                     & -3.44\%          & EfficientNet-B3                                      
\end{tabular}
    \end{minipage}
    \begin{minipage}{.49\linewidth}
        \centering
        \caption{Difference in performance between models trained using FS and VS data. This results are used to validate hypothesis \hypref{hyp:second}.}
        \label{tab:hyp2_diff}
        \begin{tabular}{@{}r|l|l@{}}

                                & \textbf{FS to VS}   & \textbf{Architecture}   \\ \midrule
\multirow{7}{*}{\textbf{R65k}}  & +0.04\%             & VGG11                   \\
                                & +0.01\%             & VGG16                   \\
                                & -0.67\%             & ResNet18                \\
                                & -0.22\%             & ResNet50                \\
                                & +2.92\%             & DenseNet121             \\
                                & -1.35\%             & EfficientNet-B0         \\
                                & -1.63\%             & EfficientNet-B3         \\ \midrule
\multirow{4}{*}{\textbf{R360k}} & +0.38\%             & VGG11                   \\
                                & -3.56\%             & ResNet18                \\
                                & -9.14\%             & EfficientNet-B0         \\
                                & -8.91\%             & EfficientNet-B3         \\                    
\end{tabular}

% VGG11 +0.04\% (equivalent)
% VGG16 +0.01\% (equivalent)
% ResNet18 -0.67\%
% ResNet50 -0.22\%
% DenseNet121 +2.92\%
% EfficienNet-B0 -1.35\%
% EfficienNet-B3 -1.63\%

% VGG11 +0.38\% (equivalent)
% ResNet18 -3.56\%
% EfficienNet-B0 -9.14\%
% EfficienNet-B3 -8.91\%

    \end{minipage}
\end{table}

% H1 results discussion
\hypref{hyp:first} is validated for 6 out of 8 comparison pairs. In these, the models trained on HR data (\ie R360k) achieve a boost in performance around 4\%. This happens for all four architectures tested (VGG11, ResNet18 and the EfficientNet variants). Noticeably, the two cases where using HR data does not yield benefits (and instead degrades around 3\%) are VS settings using the EfficientNet architectures. This phenomenon will be further studied and discussed next, when we assess the validity of \hypref{hyp:second} hypothesis.

% H2
Let us now consider the second hypothesis \textit{\hypref{hyp:second}: MAMe benefits from VS data \wrt FS data}. Since the FS property deforms the original image adding some distortion, this hypothesis aims to measure to which degree is this deformation relevant for performance on MAMe. For this purpose we compare models which have the same resolution and only differ in shape; we compare R65k-FS with R65k-VS, and R360k-FS with R360k-VS. The architectures used for the R65k comparison are all architectures listed in section \ref{sec:base_arc}, but architectures used in the R360k comparison are a subset of them due to high-memory requirements: VGG11, ResNet18, EfficientNet-B0 and EfficientNet-B3. All models are trained starting from ImageNet pre-trained models. Results are shown in Table \ref{tab:hyp1_2_results} and Table \ref{tab:hyp2_diff}.

% H2 results discussion
With regards to \textit{\hypref{hyp:second}}, results do not fully validate nor reject it. Prototypical architectures do not adequately take advantage of the VS property, either by obtaining insignificant accuracy variations (VGG11, VGG16, ResNet18-R65k and ResNet50) or even producing a negative effect on performance due to VS (ResNet18-R360k, EfficientNet-B0 and EfficientNet-B3). Only for the case of DenseNet121 on R65k data, performance increases when moving from FS to VS. Overall, results are inconclusive regarding this second hypothesis \textit{\hypref{hyp:second}}.

The experiments with regards to \textit{\hypref{hyp:second}} are affected by the padding. To process images of VS in a single batch, it is required that they all have exactly the same shape for computational purposes. Doing so without changing the shape implies the addition of padding pixels, that is, non-informative values (typically zeros) that are used to uniform batch shape. However, existing architectures do not differentiate padding pixels from image pixels, counting as noise during the training process. Remarkably, this noise increases considerably under a HR setting, where the absolute amount of padding pixels increases. In this regard, some preliminary experiments conducted on a previous version of the MAMe dataset \citep{sotiropoulos2020handling} indicate that reducing padding in a VS setting can yield to accuracy improvements between 3\% and 5\%.

% H3
As illustrated in the results of the \hypref{hyp:first}, on the MAMe dataset a gain in resolution implies a gain in performance. However, as suggested by Sandler \textit{et. al.}~\citep{sandler2019nondiscriminative}, such increase in performance may be due to an increase of the input size or due to an increase of the internal representation of the model. In their publication \citep{sandler2019nondiscriminative}, they evaluate the impact of these two factors with an experiment on the ImageNet dataset. This experiment consist on comparing information gain against resolution gain, by assessing pairs of models trained with images of same resolution but different amount of information:
\begin{itemize}
    \item Full-information images: Images that are downsampled to a target resolution and contain their corresponding amount of information.
    \item Capped-information images: Images that are, first, downsampled to a 224x224 resolution and, second, upsampled to a larger target resolution (Bilinear interpolation). Notice that such image will be of same resolution than its corresponding full-information image, but it will contain less original information.
\end{itemize}
Full-information images increase the model internal representation and the input information, while capped-information images increase the model internal representation as well, but do not increase the input information. By comparing performances using both, we can isolate the impact of the input information when increasing the image resolution.

\begin{figure}[t]
    \centering
    \includegraphics[width=0.48\textwidth]{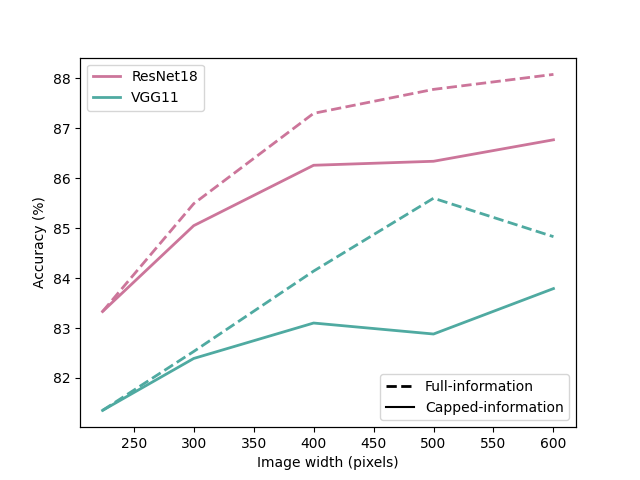}
    \caption{Results when training VGG11 and ResNet18 architectures using full-information (downsampled from original) and capped-information (upsampled from 224) images. Target resolutions used are the ones corresponding to image widths of 224, 300, 400, 500 and 600 pixels. Both settings improve with size, indicating that both resolution and information gain contribute to better performances in MAMe. Notice the experiment at 224 width is not capped information.}
    \label{fig:hyp3}
\end{figure}

To evaluate the impact of image information gain on the MAMe dataset, we formulated our third hypothesis \textit{\hypref{hyp:third}: MAMe benefits from information gain \wrt only resolution gain}. While this hypothesis is \textit{rejected} in ImageNet according to \citep{sandler2019nondiscriminative}, next we test if this is also the case for MAMe. To conduct this experiment, we train FS models on VGG11 and ResNet18 architectures using the following target resolutions: 50k, 90k, 160k, 250k and 360k pixels. These correspond to squared image widths of 224, 300, 400, 500 and 600 pixels. For each target resolution, we use full-information and capped-information images. Notice both full-information image and capped-information image are equivalent for the image width of 224 pixels. Results are shown in Figure \ref{fig:hyp3}.

Results obtained by Sandler \textit{et. al.}~\citep{sandler2019nondiscriminative} on ImageNet dataset \citep{russakovsky2015imagenet} indicate that there is no gain in performance due to information gain (see their Figure 2b). Indeed, improvement in their experiments is only caused by the use of larger model internal representations. In the MAMe dataset, this very same experiment shows that improvement occurs for both reasons. Increasing the model internal representation (capped-information) entails some consistent improvement in performance. However, performance is further boosted when also increasing the information (full-information). These results suggest the third hypothesis holds for the MAMe dataset, and further highlights the inherent differences between MAMe and ImageNet.

\section{Expert and explainability analysis of MAMe}\label{sec:exp}

The domain of artworks and heritage is defined by human technology, skill and creativity. Art experts can identify a set of visual queues useful for the characterization of art, but remains to be seen if AI models learn these same features. To analyze these features in the context of the MAMe dataset, we analyze several medium classes from an expert point of view and perform explainability experiments. We perform explainability on two models trained from scratch with two data types. On one hand, the R65k-FS is used to characterize a low resolution and fixed shape setting (LR\&FS). On the other hand, to highlight a high resolution and variable shape (HR\&VS) we introduce a new data type version: the A500-VS. This new HR\&VS version ensures a minimum size of 500 pixels per axis, preserving the original aspect ratio, as illustrated in Figure \ref{fig:data_type}. The main reason for using A500-VS instead of R360k-VS is to facilitate visualization. In this case we use the architecture most widely used for this kind of experiments, the VGG~\citep{simonyan2014very}. In our case, we use the shallow version VGG11 to handle the high-memory requirements of A500-VS. By understanding the focus of the LR\&FS and HR\&VS models, we can detect the most relevant class features according to them. These explanations allow experts to assess the consistence of the decisions made, and detect the potential existence of bias. Finally, comparison between LR\&FS and HR\&VS explanations offers an additional exploration about the impact of HR and VS properties in the MAMe dataset.

\subsection{Layer-wise relevance propagation}

In our analysis we use post-hoc interpretability \citep{mythos}: Methods used to interpret the model predictions once the model has been trained. For image classification, a widely used visual explanation are the saliency methods. These methods use saliency maps to show the features on the image that contribute to a prediction. In other words, which pixels in the input image are important for the classification task. Among this family of methods \citep{Selvaraju_2019, simonyan2013deep, springenberg2014striving, sundararajan2017axiomatic, zeiler2013visualizing}, we use Layer-wise Relevance Propagation (LRP) \citep{lrp} which has been used in different fields performing meaningful explanations \citep{10.1007/978-3-319-45886-1_28, DBLP:journals/corr/SturmBSM16, thomas2018interpretable, alex2018computational}. The LRP technique backpropagates the output prediction to the input image, by computing the contribution of each neuron \wrt the output prediction. That is, effectively mapping the relevance of an specific class into the pixels of the input image. 

\begin{figure}[t!]
    \centering
    \includegraphics[width=0.48\textwidth]{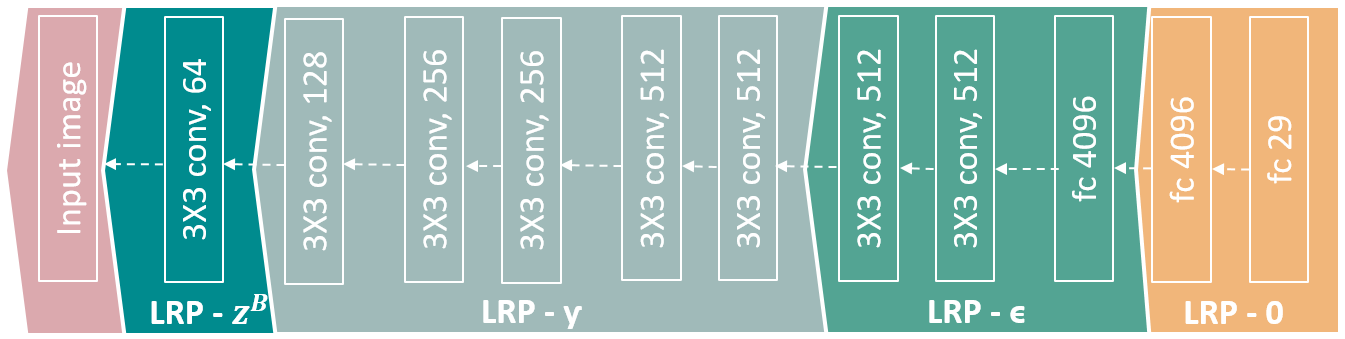}
    \caption{LRP rules applied to each layer of the VGG11 network.}
    \label{fig:lrp}
\end{figure}

Although different LRP rules have been proposed, we implement the recent Composite LRP \citep{overview}. This technique proposes to combine different propagation rules depending on the depth of the layer. Our Composite LRP makes use of LRP$ - 0$ for last layers, LRP$ - \epsilon$ ($\epsilon = 0.25$) and LRP$ - \gamma$ ($\gamma = 0.25$) for intermediate layers, and LRP$-z^B$ for the first layer of the network, as illustrated in Figure \ref{fig:lrp}.

So, given an image $I$ and a specific class $c$, the Composite LRP produces an explanation heatmap $E_{I,c}$. The color convention for this heatmap is as follows: red is used for positive contributions, while blue indicates negative contributions. That means, the red areas are considered descriptive patterns of the given class by the model. Meanwhile, the blue areas are considered typical patterns of other classes.

We perform two types of LRP analysis, one for the correctly predicted images, and another one for the incorrectly predicted images. In case of correctly predicting the medium $m$, we produce its corresponding explanation heatmap $E_{I,m}$. In this case, the red areas of the heatmap correspond to descriptive patterns of the predicted medium $m$ and blue areas to descriptive patterns of the rest of mediums. In the case of incorrectly predicted images, we computed the explanation as the difference between two heatmaps. The one associated to the real medium $r$, minus the one associated to the predicted medium $p$:
\begin{equation}
    E_{I,r,p} = E_{I,r} - E_{I,p}
\end{equation}

This difference allows to remove the contributions to the predicted class, focusing on the features that contribute to the real class. In this visualization, the red areas will be considered typical patterns of the real class but not of the predicted class, while blue areas will be considered typical patterns of other classes (most of them probably from the predicted class).

\subsection{Best and worst performances} \label{sec:exp_crc}

% Punt 1
First, let us focus on the classes that are best and worst recognized by the two models trained on the two versions of the MAMe dataset introduced before:

\begin{itemize}
    \item The model trained from scratch on images of LR\&FS (R65k-FS), using the VGG11 architecture.
    \item The model trained from scratch on images of HR\&VS (A500-VS), using the VGG11 architecture.
\end{itemize}

Among the best ones we can count \textit{Albumen photograph}, \textit{Gold} and \textit{Graphite}. In the case of \textit{Albumen photograph}, we only have one type of photographic technique in the MAMe dataset, making these images easily distinguishable from other cultural assets. The class \textit{Gold} is a similar case, since the golden color differentiates it from other metals, despite having other objects in the dataset of similar shapes. Lastly, \textit{Graphite} is a drawing technique that uses similar grey tones with metallic brightness and smooth strokes that usually end at the edge of the paper. These characteristics help avoiding confusions between \textit{Graphite} and \textit{Lithograph}, which in some cases may be similar. For these reasons these mediums are easily recognizable, not only for the LR\&FS and HR\&VS model, but also for human experts.

% Punt 2
On the other side of the spectrum we have the classes that are most poorly recognized by these two models. These are \textit{Woven fabric}, \textit{Polychromed wood}, \textit{Etching} and \textit{Silk and metal thread}. These classes are hard to predict because they belong to fine-grained groups of classes, with many common features. Following expert guidelines we identify the following fine-grained groups. These are discussed in further detail next.

\begin{itemize}
    \item \texttt{Prints}: \textit{Etching}, \textit{Engraving}, \textit{Wood engraving}, \textit{Woodcut}, \textit{Woodblock}, \textit{Lithograph}
    \item \texttt{Fabrics}: \textit{Woven fabric}, \textit{Silk and metal thread}
    \item \texttt{Paintings}: \textit{Polychromed Wood}, \textit{Oil on canvas}
\end{itemize}

\subsubsection{\texttt{Prints} group} \label{sec:prints-group}

From an expert perspective, the most complex fine-grained group is \texttt{Prints}. They are hard to differentiate because they may look very similar, despite having been printed through different procedures. Common clues used by experts for their discrimination include the definition of lines, the appearance of strokes, the homogeneity of shadows or color areas, as well as the intensity of blacks. A common feature used to identify different kinds of prints is the platemark. Platemark is the rectangular ridge created in the paper of a print by the edge of an intaglio plate. These marks can be essential for the discrimination of certain print classes: While both \textit{Engraving} or \textit{Wood engraving} have very defined lines and grid patterns, they can be told apart through platemarks since these only appear on the edges of an \textit{Engraving}. Within the same group \texttt{Prints}, \textit{Woodblocks} are distinguishable from the rest because of their oriental aesthetics. They are usually colored prints that use one block for each ink. As a result, colors sometimes overlap, and/or leave gaps in the outlines. However, this last characteristic is also found on other colored prints like \textit{Lithographs} or \textit{Woodcuts}. One last example to illustrate the complexity within \texttt{Prints} could be \textit{Etching} and \textit{Engraving}. These two techniques are very similar, having the same aforementioned platemarks and often the same grid patterns in their printed areas. In this case, experts need to appreciate the contours of the lines for differentiation. They are more vibrant and less defined in \textit{Etchings}, and they have convex edges for \textit{Engravings}.

\begin{figure}[tb]
    \centering
    \begin{tabular}{cc}
        \includegraphics[width=0.22\textwidth]{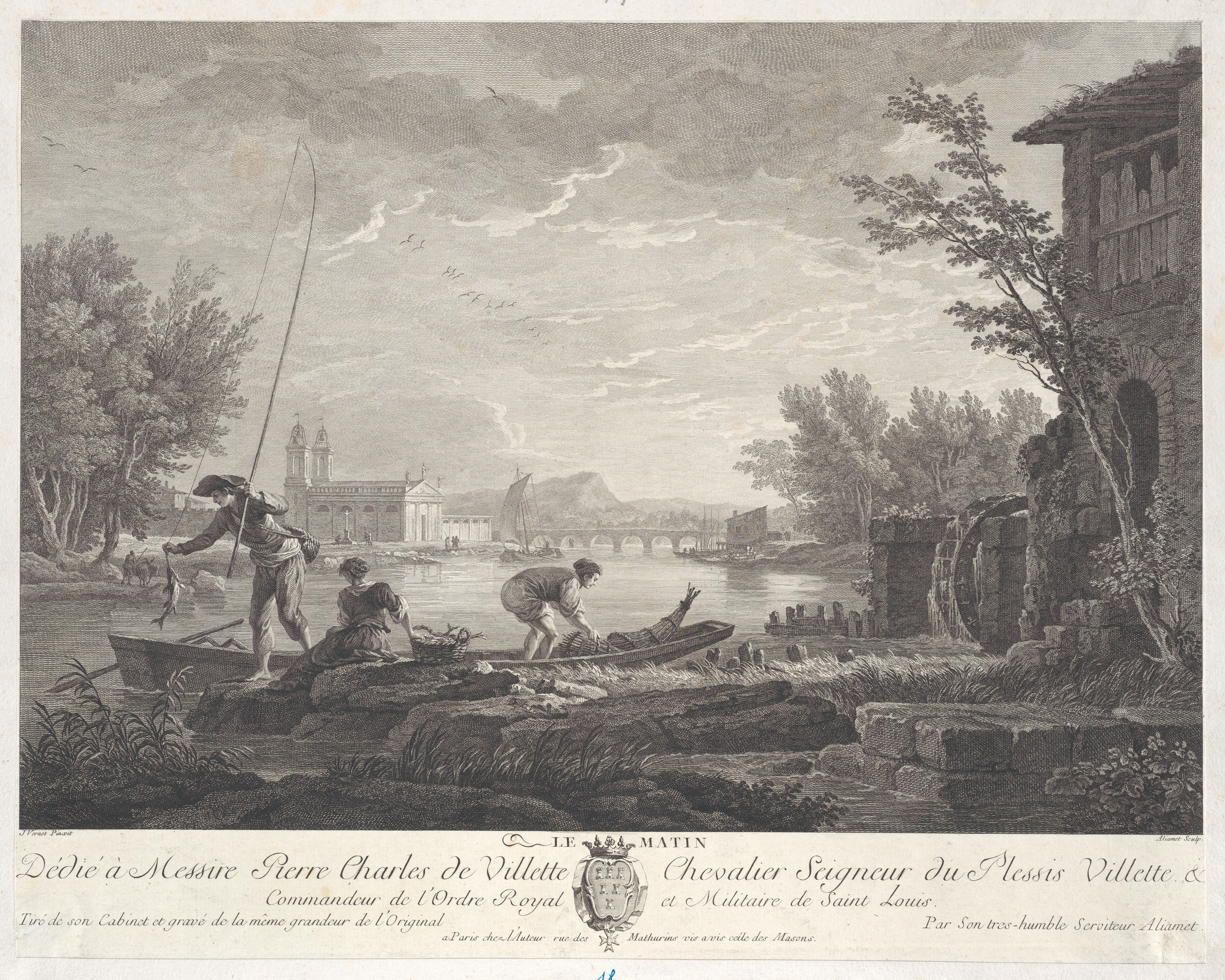} &
        \includegraphics[width=0.22\textwidth]{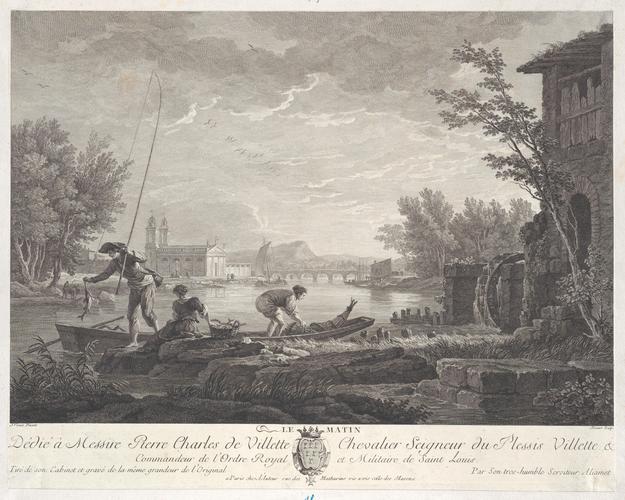}\\
        \includegraphics[width=0.22\textwidth]{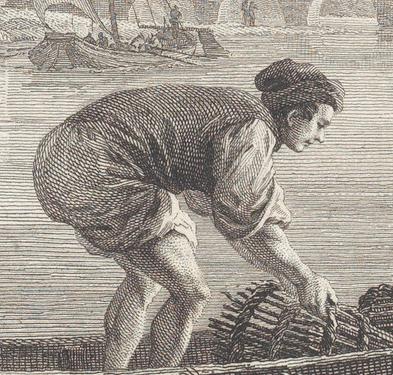} &
        \includegraphics[width=0.22\textwidth]{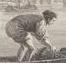}\\
    \end{tabular}   
    \caption{Example of an \textit{Engraving} artwork at its original size (left side) and HR\&VS (right side). The second row shows the same zoomed area for both images, where the grid pattern can only be perceived on the original resolution (left).\\
    \small{\textsc{Met museum: 53.600.1616}}
    }
    \label{fig:engraving-HR-resized}
    %Title: The Morning
    %Artist: After Joseph Vernet (French, Avignon 1714–1789 Paris)
    %Engraver: Jacques Aliamet (French, Abbeville 1726–1788 Paris)
    %Date: ca. 1765
    %Medium: Engraving; third state of four
    %Dimensions: Sheet (Trimmed): 13 7/16 in. × 17 in. (34.2 × 43.2 cm)
    %Classification: Prints
    %Credit Line: Harris Brisbane Dick Fund, 1953
    %Accession Number: 53.600.1616
\end{figure}

In sight of the expert knowledge, image resolution seems key to properly detect main discriminating patterns. In some cases, even our HR\&VS images seem to fall short in resolution (\eg grid patterns are lost). As an example, Figure \ref{fig:engraving-HR-resized} shows a rectangular region of an \textit{Engraving} in original resolution (left side) and in HR\&VS (right side). Zoomed area shows the central figure of the print, a fisherman. If we focus on the clothes, we can clearly perceive the characteristic grid pattern of an \textit{Engraving} in the original resolution image, but these are lost on the HR\&VS image, where the grid become a gray blur due to the interpolation when resizing the image.

\subsubsection{\texttt{Fabrics} group}
The second group of fine-grained classes is \texttt{Fabrics}. To discriminate these with total confidence it is necessary to identify the fibers using microscopy techniques. This condition motivated the aggregation of several classes within \textit{Woven fabric} (\eg linen, cotton, silk and others). Nonetheless, one particular type of woven fabric can be visually recognized without the aid of external machinery. That is \textit{Silk and metal thread}, which are clearly distinguishable from other textile fibers due to the glitter of metallic threads.

\begin{figure}[tb]
    \centering
    \begin{tabular}{cc}
        \includegraphics[height=0.18\textwidth]{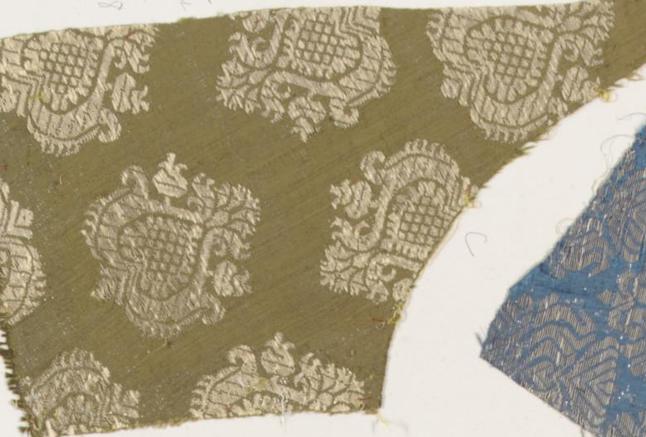} &
        \includegraphics[height=0.18\textwidth]{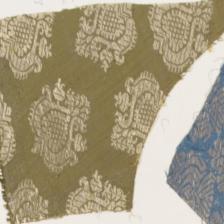}\\
    \end{tabular}   
    \caption{Example of \textit{Silk and metal thread} in HR\&VS (left) and LR\&FS (right). The brightness of the metal threads is visible in both cases.\\
    \small{\textsc{Met museum: 2002.494.278}}}
    \label{fig:silk-and-metal}
    %Title: Fragment
    %Date: 17th century
    %Culture: Italian
    %Medium: Silk, metal thread
    %Dimensions: Overall: 12 1/4 x 7 3/4 in. (31.1 x 19.7 cm)
    %Classification: Textiles-Woven
    %Credit Line: Gift of Nanette B. Kelekian, in honor of Olga Raggio, 2002
    %Accession Number: 2002.494.278
\end{figure}

\begin{figure}[tb]
    \centering
    \begin{tabular}{cc}
        \includegraphics[width=0.22\textwidth]{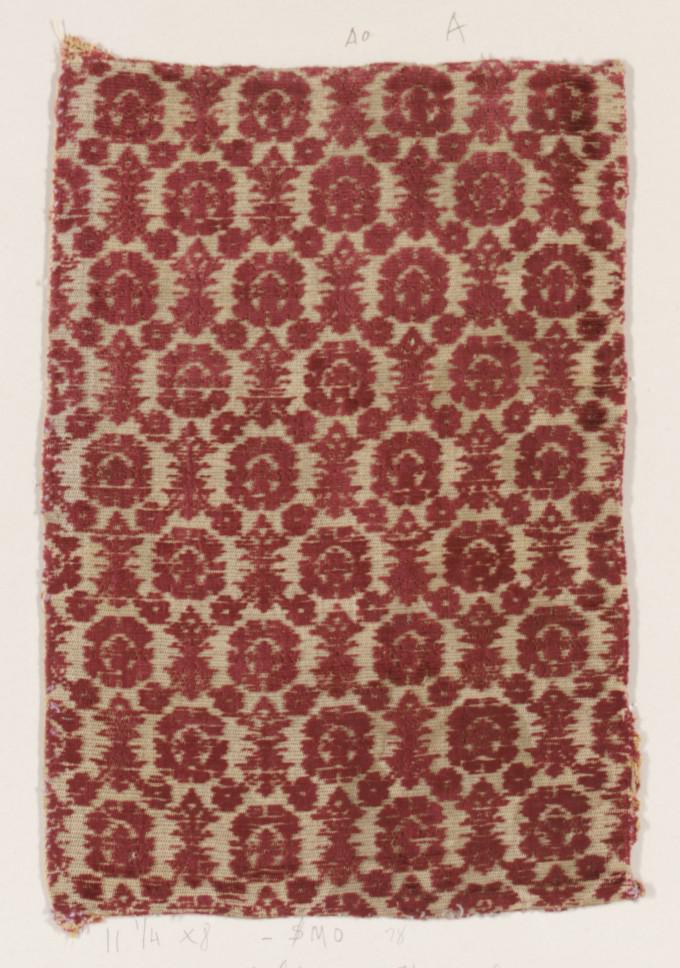} &
        \includegraphics[width=0.22\textwidth]{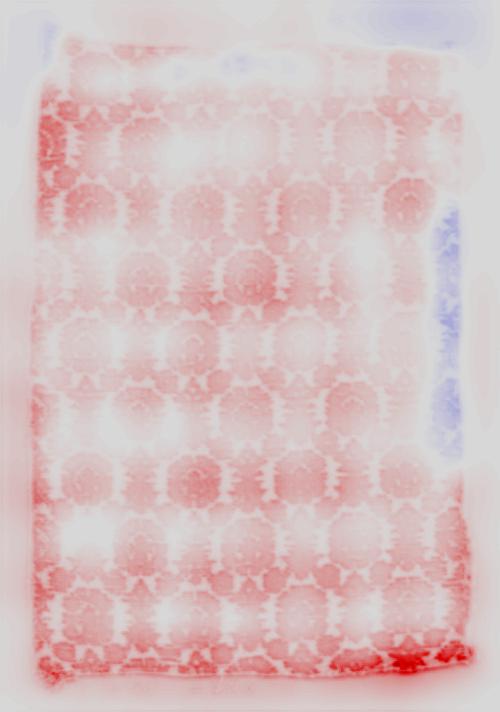}\\
    \end{tabular}   
    \caption{Example of \textit{Silk and metal thread} in HR\&VS (left) and its LRP explanation (right). The ornamental motifs (red zones) have positively contributed to the \textit{Silk and metal thread} class classification.\\
    \small{\textsc{Met museum: 2002.494.366}}}
    \label{fig:silk_and_metal_motifs}
    %Title: Fragment
    %Date: 18th century
    %Culture: Italian
    %Medium: Silk, metal thread
    %Dimensions: Overall: 11 5/8 x 8 in. (29.5 x 20.3 cm)
    %Classification: Textiles-Velvets
    %Credit Line: Gift of Nanette B. Kelekian, in honor of Olga Raggio, 2002
    %Accession Number: 2002.494.366
\end{figure}

In Figure \ref{fig:silk-and-metal}, we can see the metallic glitter in LR\&FS and HR\&VS images (more clearly on the latter). However, both models have been unable to properly discriminate these two classes. If the model does not detect this feature, it will learn other patterns for differentiating these two classes, such as ornamental motifs. However, this is not a reliable discriminatory feature and, therefore, it could be a source of error. We performed explainability experiments on several images and found cases where the model focuses on the ornamental motifs as shown in Figure \ref{fig:silk_and_metal_motifs}.

\subsubsection{\texttt{Paintings} group}
The third group of fine-grained classes is \texttt{Paintings}. This group contains two classes: \textit{Polychromed wood} and \textit{Oil on canvas}. The main reason why these classes are hard to differentiate is because \textit{Polychromed wood} contains the subclass panel paintings (\ie a painting on a flat panel made of wood), which are similar to \textit{Oil on canvas}. Both, \textit{Polychromed wood} panel paintings and \textit{Oil on canvas}, hide the support behind the paint layer, complicating the identification of the support material (fabric or wood). In this context, experts pay attention to cracks, leaks or textures that may be characteristic of the support below the paint. Nonetheless, these features may not be properly visible in a single LR\&FS or HR\&VS images.

\begin{figure}[t]
    \centering
    \begin{tabular}{cc}
        \includegraphics[width=0.22\textwidth]{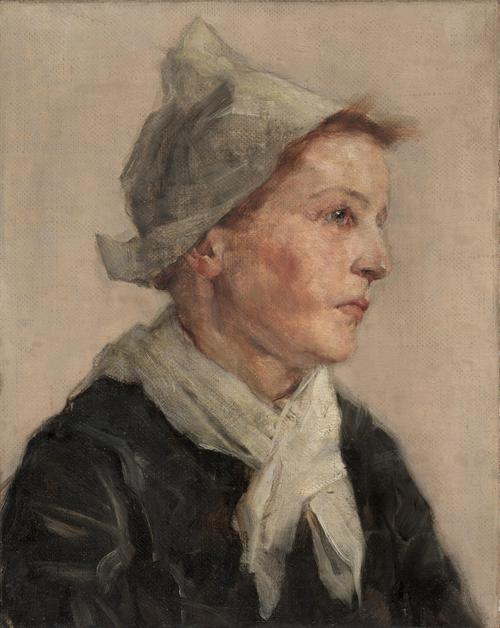} &
        \includegraphics[width=0.22\textwidth]{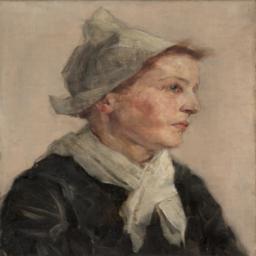}\\
        \includegraphics[width=0.22\textwidth]{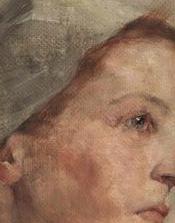} &
        \includegraphics[width=0.22\textwidth]{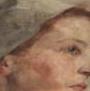}\\
    \end{tabular}
    \caption{Example of \textit{Oil on canvas} in HR\&VS (left side) and LR\&FS (right side). The second row shows the zoomed area where it is possible to perceive the canvas wave pattern in the HR\&VS but not in the LR\&FS image.\\
    \small{\textsc{Cleveland museum: 1943.324}}}
    \label{fig:polychromed-oil}
    %{{cite web|title=Head of a Woman|url=https://clevelandart.org/art/1943.324|author=Frederick Gottwald|year=late 1800s or early 1900s|access-date=24 July 2020|publisher=Cleveland Museum of Art}}
\end{figure}

There are several \textit{Oil on canvas} images that are incorrectly predicted as \textit{Polychromed wood}, both in LR\&FS and in HR\&VS. It makes sense from an expert point of view since, in several HR\&VS images, it is impossible to appreciate any detail that may suggest whether the support is wood or fabric, forcing the model to guess the class based on alternative patterns that may be misleading. For example, one of the key properties that identify an \textit{Oil on canvas} is the canvas weave pattern. Unfortunately, this seems to be visible only on a few HR\&VS images. Within this work, art experts reviewed around \~150 images where the two models failed to discriminate between \textit{Oil on canvas} and \textit{Polychromed wood}, and considered that they could only see the canvas weave pattern in approximately 5\% of the HR\&VS images. In Figure \ref{fig:polychromed-oil}, we show an example of an \textit{Oil on canvas} image where it is possible to perceive the canvas weave pattern. Although this pattern is present in the HR\&VS image but not in the LR\&FS image, both models misclassified this example, indicating that the HR\&VS model does not pay attention to this property.

\subsection{LR\&FS and HR\&VS comparison} \label{sec:exp_cri}

% Punt 3
In this section we explore the classes with greatest difference in accuracy between the two models we are exploring: the one trained on LR\&FS images and the one trained on HR\&VS images. In order, these classes are \textit{Lithograph} (+16.28\% gain by HR\&VS), \textit{Bronze} (+15.71\% gain by HR\&VS) and \textit{Engraving} (+14.85\% gain by HR\&VS). \textit{Lithograph} and \textit{Engraving} are within the \texttt{Prints} group which, as reviewed in \S\ref{sec:prints-group}, can benefit from more detailed inputs for their discrimination. The third, \textit{Bronze} is a material which can be easily differentiated by a human expert.

\begin{figure}[t]
    \centering
    \begin{tabular}{cc}
        \includegraphics[width=0.22\textwidth]{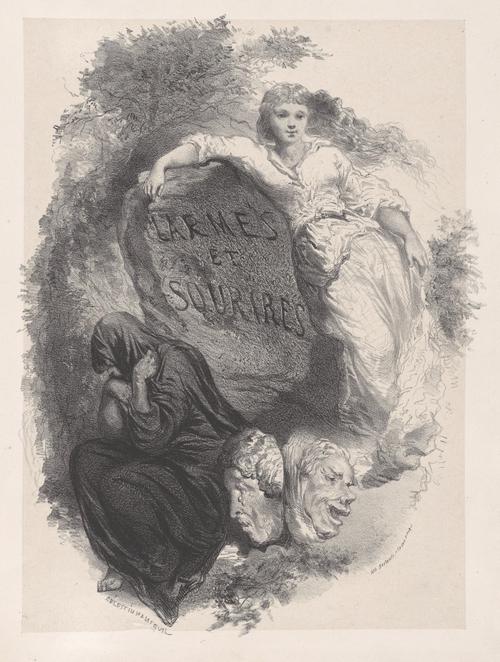} &
        \includegraphics[width=0.22\textwidth]{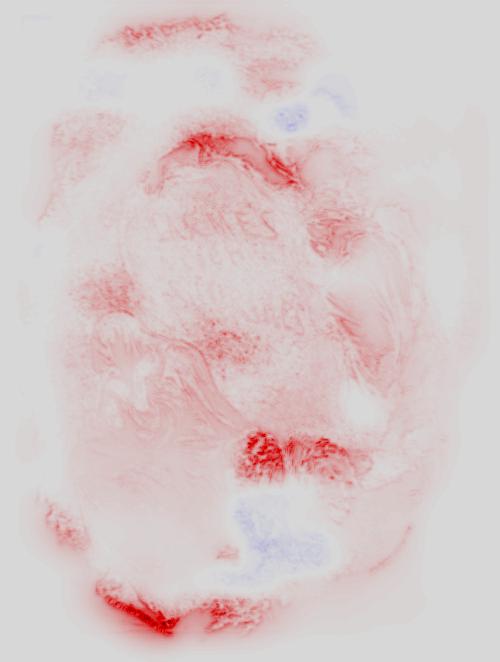} \\
        \includegraphics[width=0.22\textwidth]{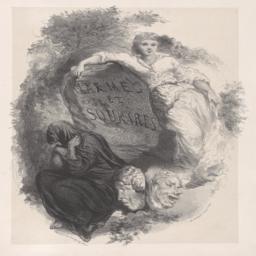} &
        \includegraphics[width=0.22\textwidth]{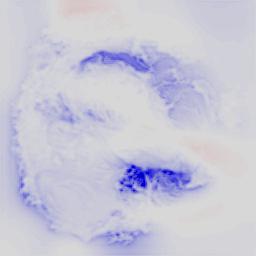} \\
        \hline \\
        \includegraphics[width=0.22\textwidth]{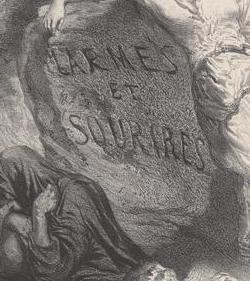} &
        \includegraphics[width=0.22\textwidth]{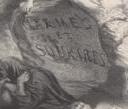} \\
    \end{tabular}   
    \caption{\textit{Lithograph} example in HR\&VS and LR\&FS. There is a top side and a bottom side divided by an horizontal black line. Top shows the image in HR\&VS and its corresponding LRP explanation. Both models focus on the general texture for their predictions, although LR\&FS mispredicts \textit{Wood engraving}. Bottom side shows a zoomed area of the print in HR\&VS (left) and LR\&FS (right). In here we can see the granular texture of the surface typical of this class in HR\&VS, but not in LR\&FS.\\
    \small{\textsc{Met museum: 49.21.53}}}
    \label{fig:812558-lithograph}
    %Title: Tears and Smiles
    %Artist: Célestin Nanteuil (French (born Italy), Rome 1813–1873 Bourron-Marlotte)
    %Printer: Bertauts, P. Cadet, Paris
    %Date: ca. 1830–60
    %Medium: Lithograph
    %Dimensions: Sheet: 13 in. × 9 15/16 in. (33 × 25.2 cm)
    %Classification: Prints
    %Credit Line: The Elisha Whittelsey Collection. The Elisha Whittelsey Fund, 1949
    %Accession Number: 49.21.53
\end{figure}

\begin{figure}[htb]
    \centering
    \begin{tabular}{cc}
        \includegraphics[width=0.22\textwidth]{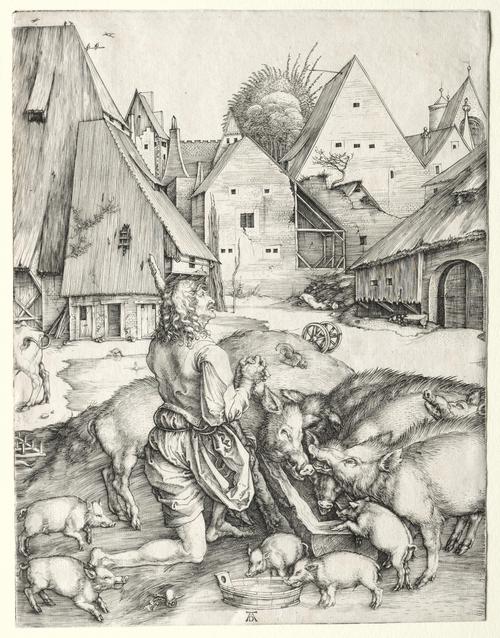} &
        \includegraphics[width=0.22\textwidth]{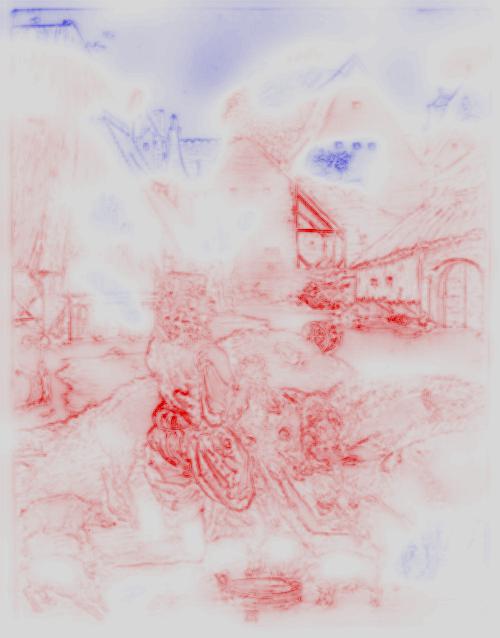} \\
        \includegraphics[width=0.22\textwidth]{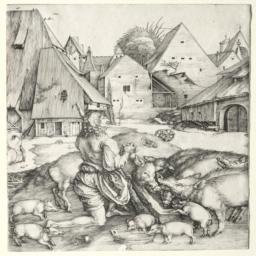} &
        \includegraphics[width=0.22\textwidth]{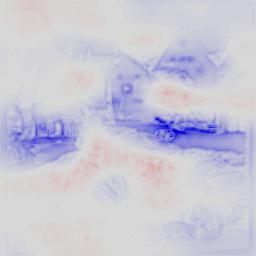} \\
    \end{tabular}   
    \caption{\textit{Engraving} example in HR\&VS and LR\&FS and its corresponding LRP explanations. Check how the contours of the figures positively contribute to the prediction of the class in HR\&VS format. LR\&FS loses most these details, and mispredicts  it as \textit{Wood engraving}.\\
    \small{\textsc{Cleveland museum: 1958.105}}}
    \label{fig:1958.105-engraving}
    %{{cite web|title=The Prodigal Son|url=https://clevelandart.org/art/1958.105|author=Albrecht Dürer|year=c. 1496|access-date=24 July 2020|publisher=Cleveland Museum of Art}}
\end{figure}

Let us start with the case of \textit{Lithograph}. Figure \ref{fig:812558-lithograph} shows a representative example of this class, illustrating both the input and the LRP for the HR\&VS and LR\&FS models. Both models focus on the overall texture of the image (the LRP relevance is spread throughout the image), but with different impacts on the prediction: it represents negative evidence for LR\&FS (which ends up in mispredicting the class \textit{Wood engraving}) but positive evidence for HR\&VS. Experts highlight the relevance of the texture of \textit{Lithographs} for their discrimination from other similar classes like \textit{Woodblock}, \textit{Hand-colored etching}, \textit{Wood engraving} or \textit{Hand-colored engraving}. \textit{Lithographs} contain a granular texture that is not present on the other classes, but this texture is only visible at a certain resolution, as shown in the zoomed tombstone at the bottom of Figure \ref{fig:812558-lithograph}. This LRP results indicate that the HR\&VS model follows a similar strategy to distinguish \textit{Lithograph} from other classes, successfully recognizing the textures from prints and properly interpreting them for the final prediction. The LR\&FS model, unable to recognize the granular texture, fails at finding relevant features towards \textit{Lithograph}.

Figure \ref{fig:1958.105-engraving} shows an example of the \textit{Engraving} class, which has been correctly predicted by the HR\&VS but not by the LR\&FS (mispredicted as \textit{Wood engraving}). The Figure contains the entire image and its corresponding LRP explanation for both, HR\&VS and LR\&FS, which target really different aspects of the print: While HR\&VS focuses on the contours of the print figures, LR\&FS does not. According to experts, these figure contours are dark areas that encode essential information for discriminating the mediums within the \texttt{Prints} group. Contours can only be properly inspected at high resolutions. Some of this information is retained in HR\&VS images, as reviewed by experts. Meanwhile LR\&FS images lose all relevant details. 

\begin{figure}[tb]
    \centering
    \begin{tabular}{cc}
        \includegraphics[height=0.13\textwidth]{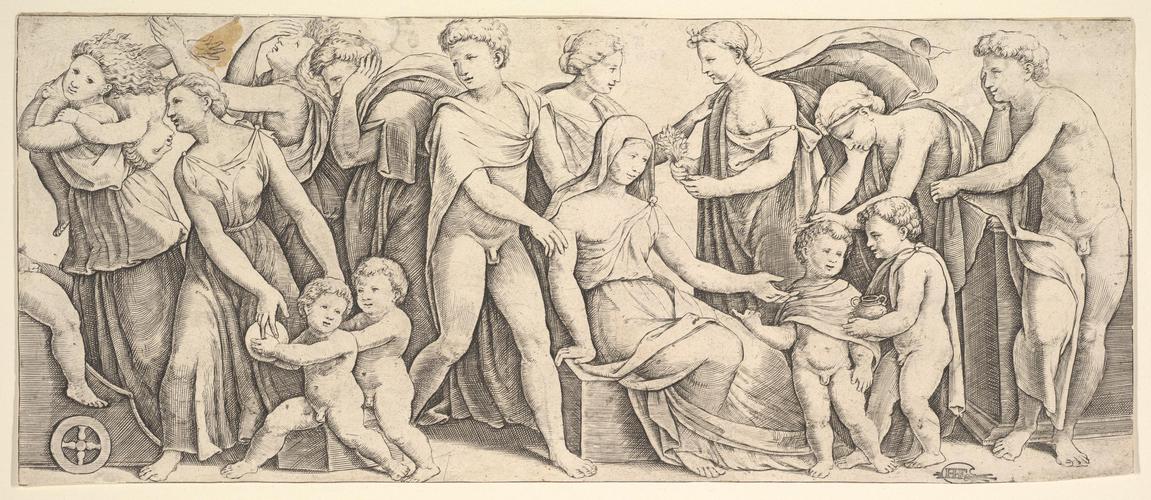} &
        \includegraphics[height=0.13\textwidth]{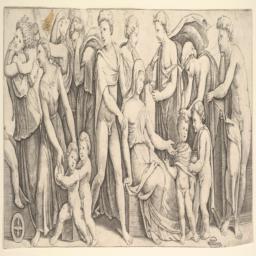}\\
        \includegraphics[height=0.16\textwidth]{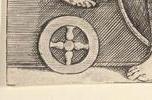} &
        \includegraphics[height=0.16\textwidth]{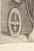}\\
    \end{tabular}   
    \caption{First row shows an \textit{Engraving} in HR\&VS and LR\&FS. Notice the deformation of the latter. Second row shows a zoomed area, to illustrate how the grid lines become blurred in the LR\&FS version.\\
    \small{\textsc{Met museum: 17.3.3169}}}
    \label{fig:396174-engraving}
    %Title: The wedding of Jason and Creusa, at left Medea takes her children
    %Artist: Master of the Die (Italian, active Rome, ca. 1530–60)
    %Date: 1530–60
    %Medium: Engraving
    %Dimensions: plate: 4 15/16 x 12 3/16 in. (12.5 x 31 cm)
    %Classification: Prints
    %Credit Line: Harris Brisbane Dick Fund, 1917
    %Accession Number: 17.3.3169
\end{figure}

As mentioned in subsection \ref{sec:prints-group}, another property to distinguish printing techniques is the grid pattern. Although in some cases it can only be perceived in the original resolution image, some HR\&VS image retain this information. However, this is always lost in the LR\&FS images. On top of that, the image distortion produced by the shape variation of LR\&FS images forces the grid lines closer in one axis (unpredictably, as it depends on the original image aspect ratio), complicating its identification. As an example of that, Figure \ref{fig:396174-engraving} shows an \textit{Engraving} image in HR\&VS and LR\&FS format, where the latter shows a great image distortion. It also shows a zoomed area, highlighting the differences in the grid pattern.

\begin{figure}[tb]
    \centering
    \begin{tabular}{cc}
        \includegraphics[height=0.14\textwidth]{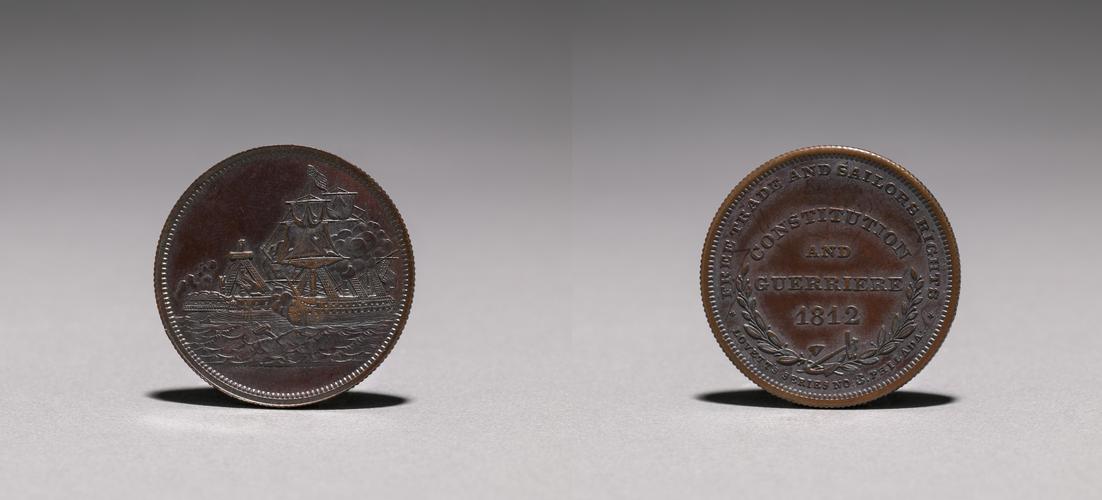} &
        \includegraphics[height=0.14\textwidth]{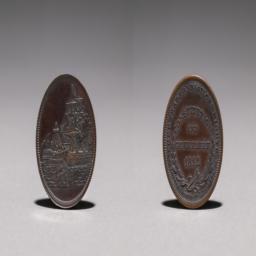} \\
        \includegraphics[height=0.14\textwidth]{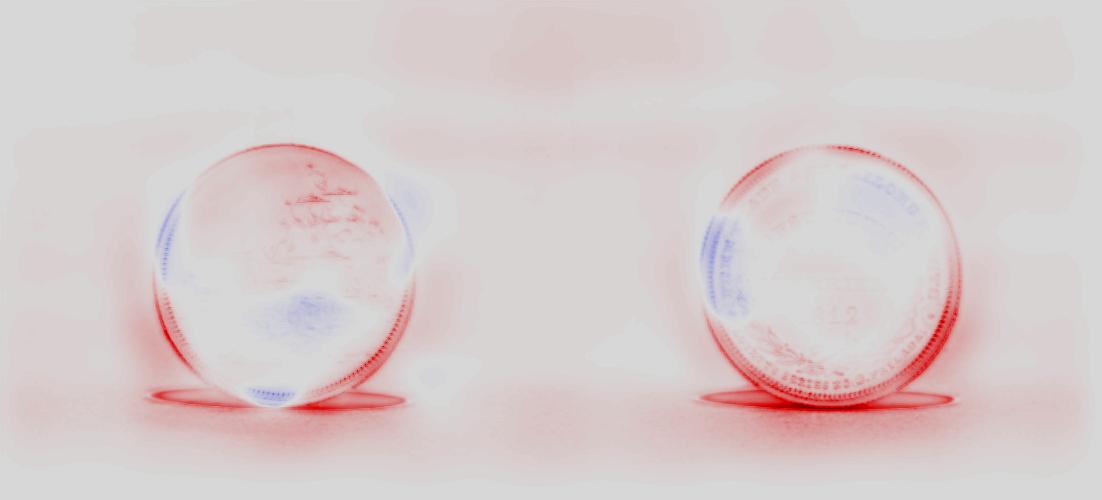} &
        \includegraphics[height=0.14\textwidth]{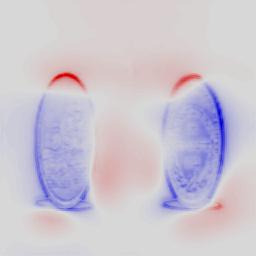} \\
    \end{tabular}   
    \caption{Example of coins within the \textit{Bronze} class in HR\&VS (left side) and LR\&FS (right side), and its corresponding LRP explanations (bottom). Shape of coins is lost in LR\&FS, which affects the prediction.\\
    \small{\textsc{Cleveland museum: 1916.1877}}}
    \label{fig:1916.1877-bronze}
    %{{cite web|title=Medal: Constitution and Guerriere, 1812|url=https://clevelandart.org/art/1916.1877|year=1812|access-date=24 July 2020|publisher=Cleveland Museum of Art}}
\end{figure}

The last case we consider in this section is the third class with the biggest difference in performance. This is the \textit{Bronze} class, which includes a great variety of objects (\eg sculptures, ornaments), but specially coins. One of the main reasons why there are so many coins inside the \textit{Bronze} class is that, historically, \textit{Bronze} has been a usual alloy used to mint coins. One of the main characteristics of a coin is its circular shape. However, this property is lost when deforming the image due to the uniformization of aspect ratio inherent to LR\&FS inputs. The lack of a uniform shape of coins has a negative impact on their recognition, which is not found on the HR\&VS model. A clear example of this can be observed in Figure \ref{fig:1916.1877-bronze}. The corresponding LRP explanations show, on one side, the positive impact of the rounded coin contour for the HR\&VS image and, on the other side, the negative impact of the deformed coin for the prediction of the LR\&FS image. This particular LR\&FS example is mispredicted with \textit{Steel}, which makes sense from an expert point of view because the model must focus on the detection of the material, as it can not rely on the shape of the coin for the prediction. Indeed, classes like \textit{Steel} and \textit{Iron} are among the most frequent confusions for \textit{Bronze}. As as result, \textit{Bronze} is significantly better predicted by the HR\&VS, with a 15.7\% increase in accuracy with respect to LR\&FS.

\begin{figure}[tb]
    \centering
    \begin{tabular}{cc}
        \includegraphics[height=0.22\textwidth]{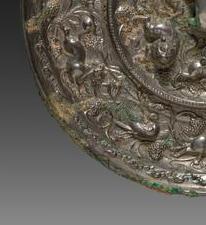} &
        \includegraphics[height=0.22\textwidth]{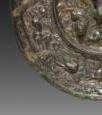} \\
    \end{tabular}   
    \caption{Zoom in of a \textit{Bronze} artwork in HR\&VS (left) and in LR\&FS (right) respectively. Notice how the corrosion and patinas are easier to appreciate in HR\&VS.\\
    \small{\textsc{Cleveland museum: 1926.248}}}
    \label{fig:1926.248-bronze}
    %{{cite web|title=Mirror|url=https://clevelandart.org/art/1926.248|year=618-907|access-date=24 July 2020|publisher=Cleveland Museum of Art}}
\end{figure}

Another example is shown in Figure \ref{fig:1926.248-bronze}, where we can see the characteristic corrosion and patinas of \textit{Bronze}. This corrosion or green patinas on the surface comes from the oxidation of copper, which is one of the main components of the \textit{Bronze} alloy. Experts underline that these properties make quite easy to recognize the class. While these are perfectly visible in HR\&VS images, they become hard to perceive in the LR\&FS images.

\section{Conclusions}\label{sec:c}

% More gain in performance due to resolution gain than depth

In this paper, we introduce the MAMe dataset, a novel challenge for the prediction of artwork mediums based on its visual appearance. The images of the dataset come from three different museums for a total of 37,407 images. Museums do not share a common scheme for labeling mediums, which required intensive work by art experts for its homogenization. For producing the dataset, we leverage technical requirements (sample size, balance, image resolution, \etc) and domain requirements (visual coherency, taxonomical properties, \etc). At the end, the MAMe is composed by 29 classes of mediums, each containing at least 850 images (always 700 for training) of high resolution (at least 500 pixels in the smaller axis) and variable shape.

In comparison with commonly available datasets, the MAMe provides a significantly larger distribution of high resolution and variable-shaped images. These properties are of relevance for future applications in domains such as medicine or autonomous driving; domains where attention to detail, understanding the overall structure and avoiding image pattern deformation/loss is crucial. Recognizing a lack of focus on these topics by the AI community, MAMe provides a good testing environment for new research ideas in the field.

% HR better than LR
% VS better than FS
% Information gain improves performance
Baselines and hypothesis results provide several conclusions. Regarding baselines, results presented in Table \ref{tab:baseline_results} show the capability of such models to solve the task proposed by the MAMe dataset up to certain degree, with a top performance of 88.95\% accuracy achieved by EfficientNet-B3 architecture using R360k-FS data format. Regarding hypothesis evaluation, results shown in Table \ref{tab:hyp1_2_results} and Figure \ref{fig:hyp3} support \hypref{hyp:first} and \hypref{hyp:third} hypotheses but not \hypref{hyp:second}. We conclude from our first hypothesis \hypref{hyp:first} that performance on the MAMe task increases when using images of high resolution over standard low resolution ones. Furthermore, based on the validated third hypothesis \hypref{hyp:third}, we see that this performance gain comes not only from larger image resolution but also from an increase of the image information (unlike ImageNet \citep{russakovsky2015imagenet}). In contrast, results on the \hypref{hyp:second} hypothesis do not validate it. We consider prototypical architectures to be specifically designed for FS data, hence not taking proper advantage of the VS property. Moreover, the current way of handling VS introduce padding on batching, increasing the amount of noise during training, specially in high resolution settings (\ie R360k in our case). We consider that there is room for improvement in this area based on studies in a previous version of the MAMe dataset, where padding reduction on VS models provides performance improvements of 3\% to 5\%  \citep{sotiropoulos2020handling}.

Lastly, we perform an explainability and expert analysis to further understand the differences when training models with either, low resolution and fixed-shape (R65k-FS) images or high resolution and variable-shape (A500-VS). The results of these analysis allow us to assess how the models we explored fail to discriminate between certain classes due to a lack of resolution. In several cases we found that even the A500-VS resolution is insufficient to perceive the patterns that experts would pay attention to. This forces the models to learn on alternative patterns that may not generalize well.

% \dgnote{final paragraf: mame is different than others. mame benefits of HR. VS needs more research.}
Overall, the MAMe dataset constitutes a large scale challenge which benefits from the use of HR. Benefit from using HR images does not only come from bigger internal representation of the models, but also from an increase of the image information. This is particularly characteristic for MAMe, since differs from the prototypical and well-known ImageNet dataset. Further research is needed to efficiently handle VS data, and MAMe dataset serves as a good candidate for such use case.

\section*{Acknowledgments}
This work is partially supported by the Intel-BSC Exascale Lab agreement, by the Spanish Government through Programa Severo Ochoa (SEV-2015-0493), by the Spanish Ministry of Science and Technology through TIN2015-65316-P project, by the Generalitat de Catalunya (contracts 2017-SGR-1414) and by the Secretaria d’Universitats i Recerca of the Generalitat de Catalunya under the 
 Industrial Doctorate Grant DI 2018-100. Authors would like to thank the support and assessment of the Conservació-Restauració del Patrimoni group (2017-SGR-1151).

\bibliographystyle{unsrtnat}
\bibliography{mame}   % name your BibTeX data base

\end{document}